\documentclass[lettersize,journal]{IEEEtran}
\usepackage{amsmath,amsfonts}
\usepackage{algorithmic}
\usepackage{algorithm}
\usepackage{array}
\usepackage[caption=false,font=normalsize,labelfont=sf,textfont=sf]{subfig}
\usepackage{textcomp}
\usepackage{stfloats}
\usepackage{url}
\usepackage{verbatim}
\usepackage{graphicx}
\usepackage{cite}
\usepackage{multirow}
\usepackage{makecell}
\usepackage{tabularx}
\usepackage{amssymb}
\usepackage{pifont}
\usepackage{booktabs}
\newcommand{\midtilde}{\raisebox{0.5ex}{\texttildelow}} 

\usepackage{xcolor}

\begin{document}

\title{SparseMap: A Sparse Tensor Accelerator Framework Based on Evolution Strategy }


\author{Boran Zhao$^{1}$, Haiming Zhai$^{2}$, Zihang Yuan$^{1}$, Hetian Liu$^{1}$, Tian Xia$^{2}$, Wenzhe Zhao$^{2}$ \\ and Pengju Ren$^{2}$~\IEEEmembership{Member,~IEEE}   \vspace{-7mm}

        \thanks{- $^{1}$Boran Zhao, Zihang Yuan and Hetian Liu are with the School of Software Engineering, Xi'an Jiaotong University, Xi'an, Shaanxi, China.}
        \thanks{- $^{2}$All other authors are with the National Key Laboratory of Human-Machine Hybrid Augmented Intelligence, National Engineering Research Center for Visual Information and Applications, and Institute of Artificial Intelligence and Robotics, Xi'an Jiaotong University, Xi’an, Shaanxi, China.}

		\thanks{- E-mail: pengjuren@xjtu.edu.cn (Corresponding Author).}
	}

\markboth{Journal of \LaTeX\ Class Files,~Vol.~14, No.~8, August~2021}%
{Shell \MakeLowercase{\textit{et al.}}: A Sample Article Using IEEEtran.cls for IEEE Journals}


\maketitle

\begin{abstract}

The growing demand for sparse tensor algebra (SpTA) in machine learning and big data has driven the development of various sparse tensor accelerators. However, most existing manually designed accelerators are limited to specific scenarios, and it's time-consuming and challenging to adjust a large number of design factors when scenarios change. Therefore, automating the design of SpTA accelerators is crucial. Nevertheless, previous works focus solely on either \textit{mapping} (i.e., tiling communication and computation in space and time) or \textit{sparse strategy} (i.e., bypassing zero elements for efficiency), leading to suboptimal designs due to the lack of comprehensive consideration of both. A unified framework that jointly optimizes both is urgently needed. However, integrating mapping and sparse strategies leads to a combinatorial explosion in the design space(e.g., as large as $O(10^{41})$ for the workload $P_{32 \times 64} \times Q_{64 \times 48} = Z_{32 \times 48}$). This vast search space renders most conventional optimization methods (e.g., particle swarm optimization, reinforcement learning and Monte Carlo tree search) inefficient. 


To address this challenge, we propose an evolution strategy-based sparse tensor accelerator optimization framework, called SparseMap. SparseMap constructing a more comprehensive design space with the consideration of both mapping and sparse strategy. We introduce a series of enhancements to genetic encoding and evolutionary operators, enabling SparseMap to efficiently explore the vast and diverse design space. We quantitatively compare SparseMap with prior works and classical optimization methods, demonstrating that SparseMap consistently finds superior solutions.


\end{abstract}


\section{Introduction}

\IEEEPARstart{S}{parse} tensor algebra (SpTA) is widely used in various fields, such as neural networks\cite{sparse_net_1,sparse_net_2}, recommendation systems\cite{recommend_1,recommend_2}, and scientific simulations\cite{sci_simu}. General parallel hardware like GPGPUs often exhibit poor performance in many SpTA tasks due to insufficient utilization of hardware resources \cite{gpgpu1,2016Eyeriss}. As a result, an increasing number of sparse tensor accelerators have been proposed\cite{2016Eyeriss,cnvlutin,eyerissV2,extensor,matraptor,gospa,outerspace,scnn,sigma,sparten,sparsity_recon}. These acclerators use different mappings to promote data reuse and deploy various sparse strategies to take advantage of the abundance of zero elements in sparse tensors to reduce unnecessary data storage, movement and computation \cite{2021Sparseloop}. However, achieving high utilization with these accelerators often demands sophisticated manual design, which is both time-consuming and inefficient. This challenge is further exacerbated by the multitude of design parameters involved, making it difficult to identify the optimal solution for specific workloads, particularly when adapting to new sparse workloads.


\begin{figure}[t]
    \centering
    \includegraphics[width=0.9\linewidth]{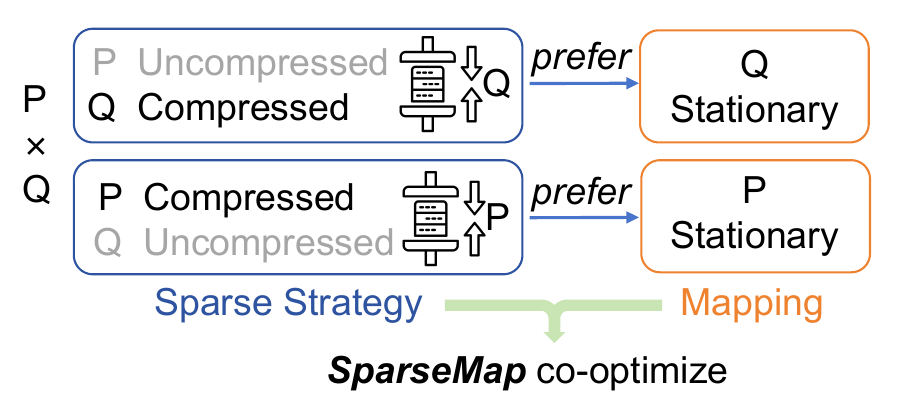} \vspace{-4mm}
    \caption{Different sparse strategies prefer different mappings. \textcolor{black}{For example, NVDLA\cite{nvdla}, STC\cite{stc_nvidiaA100}, and WOS\cite{wos} fall into the first category. By contrast, SCNN\cite{scnn}, ExTensor\cite{extensor} and SpArch\cite{sparch} belong to the second category.} Therefore, sparse strategy and mapping need to be co-optimized. \textcolor{black}{In actual, the selection of sparse strategy and mapping is governed by many factors, including tensor size, density ratio and the available on-chip SRAM etc.\cite{2021Sparseloop,2023REMAP,2020GAMMA,kao2020confuciux}. } }
    \label{fig:fig1} \vspace{-4mm}
\end{figure}

\begin{table}[t]
\centering
\caption{Comparison of existing accelerator optimization frameworks with our work.}
\label{tab_acc_framework}
\resizebox{0.4\textwidth}{!}{
    \begin{tabular}{|c|c|c|} 
    \hline
                           & \makecell{Mapping\\Exploration} & \makecell{Sparse-Strategy\\Exploration} \\ \hline
    
    REMAP\cite{2023REMAP}        & \checkmark            & $\times$                 \\ \hline
    Sparseloop Mapper\cite{2021Sparseloop}\    & \checkmark            & $\times$                 \\ \hline
    GAMMA\cite{2020GAMMA}        & \checkmark            & $\times$                \\ \hline
    Ichrak\cite{ocf1}        & $\times$           &\checkmark                \\ \hline         
    CDPU \cite{ocf2_cdpu}       & $\times$            & \checkmark               \\ \hline  
    SAGE \cite{ocf3_extending}       & $\times$            & \checkmark               \\ \hline 
    Our work     & \checkmark           & \checkmark                 \\ \hline
    \end{tabular}}  \vspace{-4mm}
\end{table}

\begin{figure}
    \centering
    \includegraphics[width=0.9\linewidth]{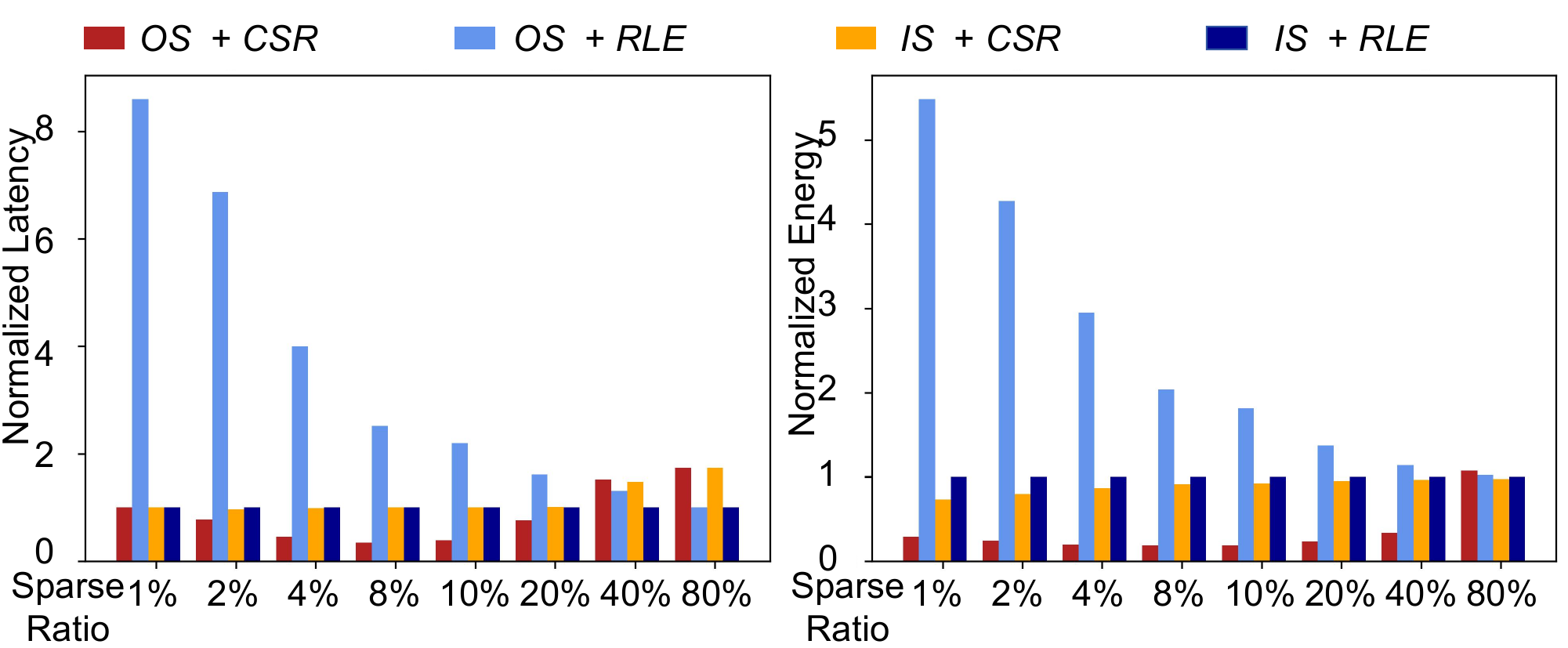}
    \caption{Impact of different mappings ($OS$ represents output stationary and $IS$ indicates input stationary) and sparse strategies ($CSR$ represents compressed sparse row and $RLE$ indicates run length encoding) on normalized latency and energy. The left plot shows normalized latency, while the right plot depicts normalized energy across varying sparsity levels. It can be observed that no single sparse strategy is optimal for all workloads and mappings, and similarly, no single mapping is optimal for all workloads and sparse strategies.}  \vspace{-4mm}
    \label{fig:diff_m_ss}
\end{figure}

To solve this problem, as shown in Table~\ref{tab_acc_framework}, many previous studies\cite{2020GAMMA,2023REMAP,2021Sparseloop,mapping_dse,ocf1,ocf2_cdpu,ocf3_extending} have been presented, which can be broadly categorized into two types. The first type employs carefully crafted search algorithms to efficiently identify mappings with the lowest latency or energy consumption for specific workloads\cite{2020GAMMA,2023REMAP,mapping_dse}, but these approaches fail to consider sparse strategies. As a result, they are only applicable to the design of dense tensor accelerators and cannot guide sparse tensor accelerator design. The second type\cite{ocf1,ocf2_cdpu,ocf3_extending} focuses on sparse strategies, such as finding the optimal tensor compression format, but they ignore mapping optimization. However, as illustrated in Fig.\ref{fig:fig1}, different sparse strategies prefer different mappings. Moreover, both prior research \cite{2021Sparseloop,ocf3_extending} and our experiments, as illustrated in Fig.\ref{fig:diff_m_ss}, indicate that no single sparse strategy is universally optimal across all mappings ($OS$ and $IS$  represent different mappings, while $CSR$ and $RLE$ denote different compression formats). Consequently, restricting exploration to sparse strategies alone while disregarding or fixing mappings often leads to suboptimal solutions.

Based on the above analysis, the design space of sparse tensor accelerators should simultaneously consider both mapping and sparse strategy. However, the combinatorial design space exhibits two critical characteristics: 1) the combinatorial explosion results in an extremely large design space (e.g., for the workload $P_{32 \times 64} \times Q_{64 \times 48} = Z_{32 \times 48}$, it can reach up to ($O(10^{41})$); 2) the design space contains a significant number of invalid design points due to mismatches between mapping and sparsity strategies, leading to resource over-utilization beyond hardware constraints. \textcolor{black}{As illustrated in Fig.~\ref{fig:fig1}, when the sparse strategy keeps 
 \( P \) uncompressed and the mapping fixes \( P \) as stationary, \( P \) must reside on-chip; under limited SRAM, this requirement can exceed the available capacity.}  These characteristics render many optimization methods inefficient (e.g., Particle Swarm Optimization (PSO), Reinforcement Learning (RL) and Monte Carlo Tree Search (MCTS)). For instance, reinforcement learning algorithms suffer from the \textit{sparse reward} problem \cite{sparse_reward1,sparse_reward2} due to the abundance of invalid solutions, making convergence unattainable.
To address this challenge, we propose a novel evolution strategy-based framework, called SpareMap. The key novelties of SparseMap are: 1) a unique genetic encoding that significantly reduces the search space while maintaining completeness; 2) customized evolutionary operators and initialization that maintain population diversity while mitigating the impact of invalid individuals.

Experimental results demonstrate that SparseMap achieves significant improvements over existing approaches. Compared to SAGE-like solutions \cite{ocf3_extending}, \textcolor{black}{SparseMap achieves an average Energy-Delay Product (EDP) reduction of $26.8\times$ on the edge platform, $19.2\times$ on the mobile platform, and $171.4\times$ on the cloud platform. Additionally, SparseMap outperforms Sparseloop Mapper \cite{2021Sparseloop}, achieving an average EDP reduction of $8.8\times$, $4.5\times$, and $158.9\times$ on the edge, mobile, and cloud platforms, respectively.} These results highlight the effectiveness of SparseMap in navigating the complex design space of sparse tensor accelerators and optimizing performance across diverse hardware resources.

The contributions of this paper are as follows:
\begin{itemize}
\item \textcolor{black}{\textbf{Essential-Factor Design Space.}} \textcolor{black}{SparseMap constructs a enormous design space for sparse tensor accelerators, encompassing both mapping and sparse strategy. Considering both design aspects significantly enhances the likelihood of finding the optimal design.} 

\item \textbf{Unique Genetic Encoding and Decoding Scheme.} We propose a novel encoding and decoding mechanism that represents a sparse tensor accelerator design as a genome (a 1D array), effectively reducing the search space while preserving all valid designs.


\item \textbf{Customized Evolutionary Operators and Initialization.} 
We introduce \textit{annealing mutation} and \textit{sensitivity-aware crossover} operators to better balance global and local search. Additionally, we propose a \textit{high-sensitivity hypercube initialization} that enhances population diversity while effectively mitigating the problem of invalid individuals. These advancements enable the successful application of evolution strategy to the SpTA accelerator design problem.

\end{itemize}

\section{Background}

\begin{figure}[t]
    \centering
    \includegraphics[width=1\linewidth]{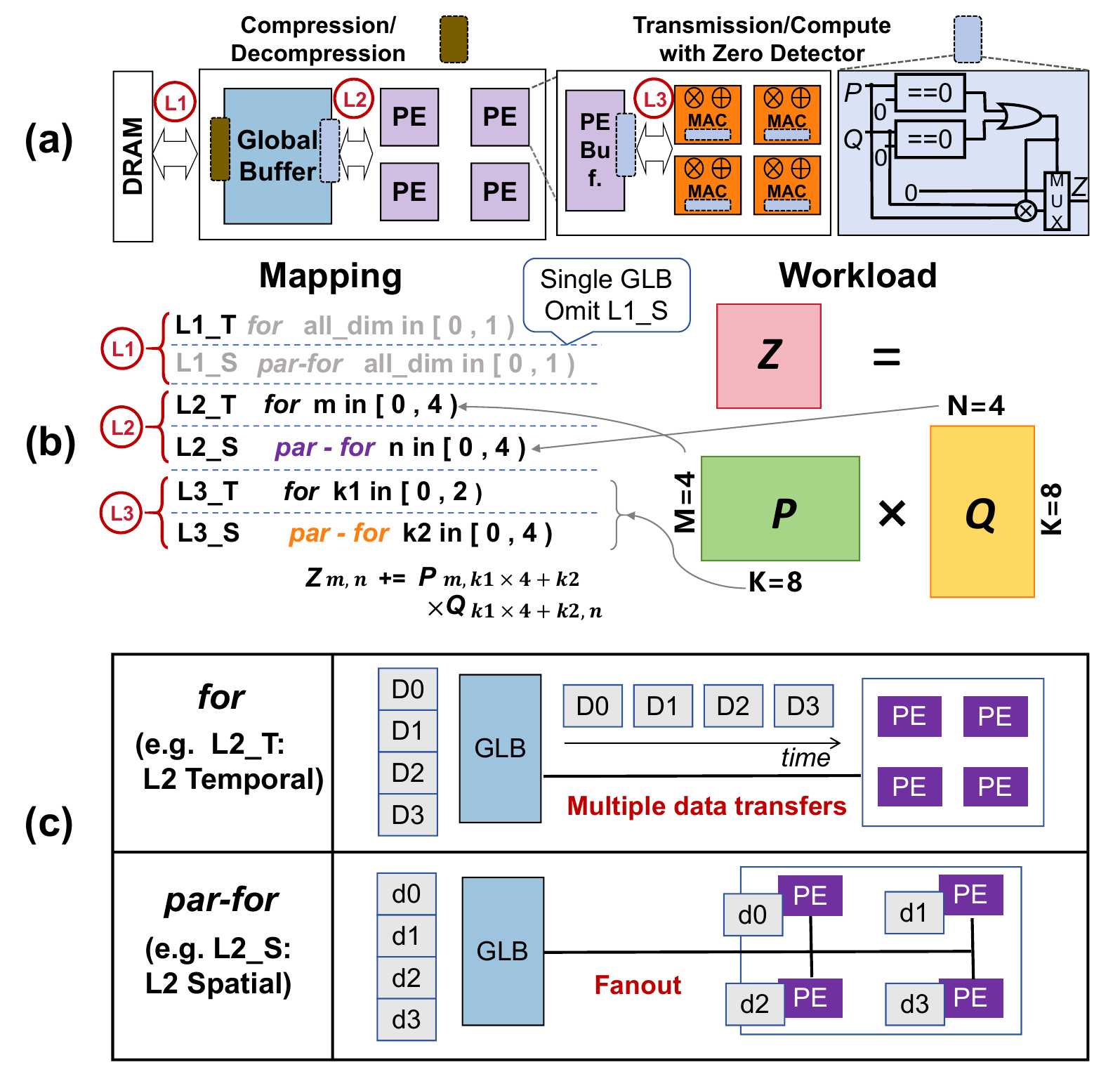}
    \caption{(a) A basic sparse tensor accelerator architecture with 3-level memory. It has an off-chip DRAM, an Global Buffer (GLB) and 4 PEs. Each PE has a  buffer and 4 MACs. (b) Illustration of mapping at different levels of the accelerator architecture. (c) At the L2 mapping level, the loop for $n$ in $[0,4)$ represents splitting the data related to the $N$ dimension in the GLB into four parts ($D0$, $D1$, $D2$, $D3$), which are then transmitted sequentially to the PE array over time. The inner loop par-for $m$ in $[0,4)$ indicates that the data related to the $M$ dimension is split into four parts, which are assigned separately to the four PEs simultaneously. Notably, due to the single instance of the GLB in this accelerator architecture, no spatial mapping exists at the L1 mapping level (from DRAM to GLB).}
    \label{fig_mapping_1}  \vspace{-4mm}
\end{figure}


\subsection{Architecture of Sparse Tensor Accelerator}

Sparse tensor accelerators typically consist of an off-chip DRAM, an on-chip Global Buffer (GLB) , and an array of Processing Elements (PEs) \cite{2016Eyeriss,eyerissV2,extensor,cnvlutin,scnn}, as shown in Fig.~\ref{fig_mapping_1}(a). Each PE has many Multiply-Accumulate Units (MACs) to compute partial sums, and local buffers (called PE Buffer in this paper) to store input tensors and partial sums. GLB is used to prefetch input tensors from DRAM for the next tile of computation that will be mapped over the PEs. In addition, sparse tensor accelerators also include specially designed logic circuits for eliminating computations and memory accesses related to zeros.


\subsection{Mapping Scheme}

Mapping refers to a set of rules that define how tasks are distributed across computational units and memory hierarchies, as well as the order in which data is accessed and processed. For example, as illustrated in Fig.~\ref{fig_mapping_1} (b), a mapping for sparse matrix multiplication ($M \times K$ and $K \times N$, where $M=4$, $K=8$ and $N=8$) are applied on the sparse tensor accelerator described in (a). The mapping can be represented by a multi-level nested for-loop. For the same SpTA workload, it can be mapped onto hardware in various ways, making the mappings diverse. (e.g., Eyeriss\cite{2016Eyeriss} uses a row-stationary mapping, DSTC\cite{dstc} uses an output-stationary mapping). Since the accelerator architecture has three memory levels, each mapping can be divided into L1, L2, and L3 from outer to inner loops, which define the movement and allocation of data between DRAM and GLB, GLB and PE, and PE buffer and MAC, respectively. As shown in Fig.~\ref{fig_mapping_1} (c), L1, L2, and L3 include \textit{for} (denoted as temporal tiling, abbreviated as ``\_T") and \textit{par-for} (denoted as spatial tiling, abbreviated as ``\_S"), the \textit{for} means scheduling the workload onto hardware within multiple cycles, the \textit{par-for} stands for allocating workload onto multiple process elements, simultaneously, for parallel computation.

\subsection{Sparse Strategy}

Sparse strategy defines how an accelerator exploits the sparsity of tensors to minimize unnecessary data storage, data movement, and computations, thereby enhancing overall efficiency. It can be broadly divided into two components: compression format and Skipping/Gating (S/G) mechanism \cite{2021Sparseloop,ocf3_extending}.


\begin{itemize}
    \item Compression format refers to the method used to encode the locations of nonzero elements in a tensor. For instance, Eyeriss V2 \cite{eyerissV2} employs a \textit{compressed sparse column (CSC)} format, while DSTC \cite{dstc} uses a two-level \textit{Bitmask} scheme. These formats optimize data representation to reduce storage and access overheads.
    
    \item S/G mechanism defines how hardware avoids unnecessary operations on zero elements to improve computational efficiency: \ding{182} \textit{Gating mechanism}: This approach, used by Eyeriss \cite{2016Eyeriss}, ensures that when an element in the input tensor is zero, the corresponding circuit remains idle, avoiding energy consumption associated with accessing the corresponding tensor element. \ding{183} \textit{Skipping mechanism}: Deployed by accelerators like Eyeriss V2 \cite{eyerissV2} and SCNN \cite{scnn}, this method skips over zero elements in the input tensor, directly locating the next nonzero element and accessing the corresponding tensor element. This saves both energy and computational cycles by reducing unnecessary operations.

\end{itemize}




\section{Design Space and Challenge}

\begin{figure}
    \centering
    \includegraphics[width=0.9\linewidth]{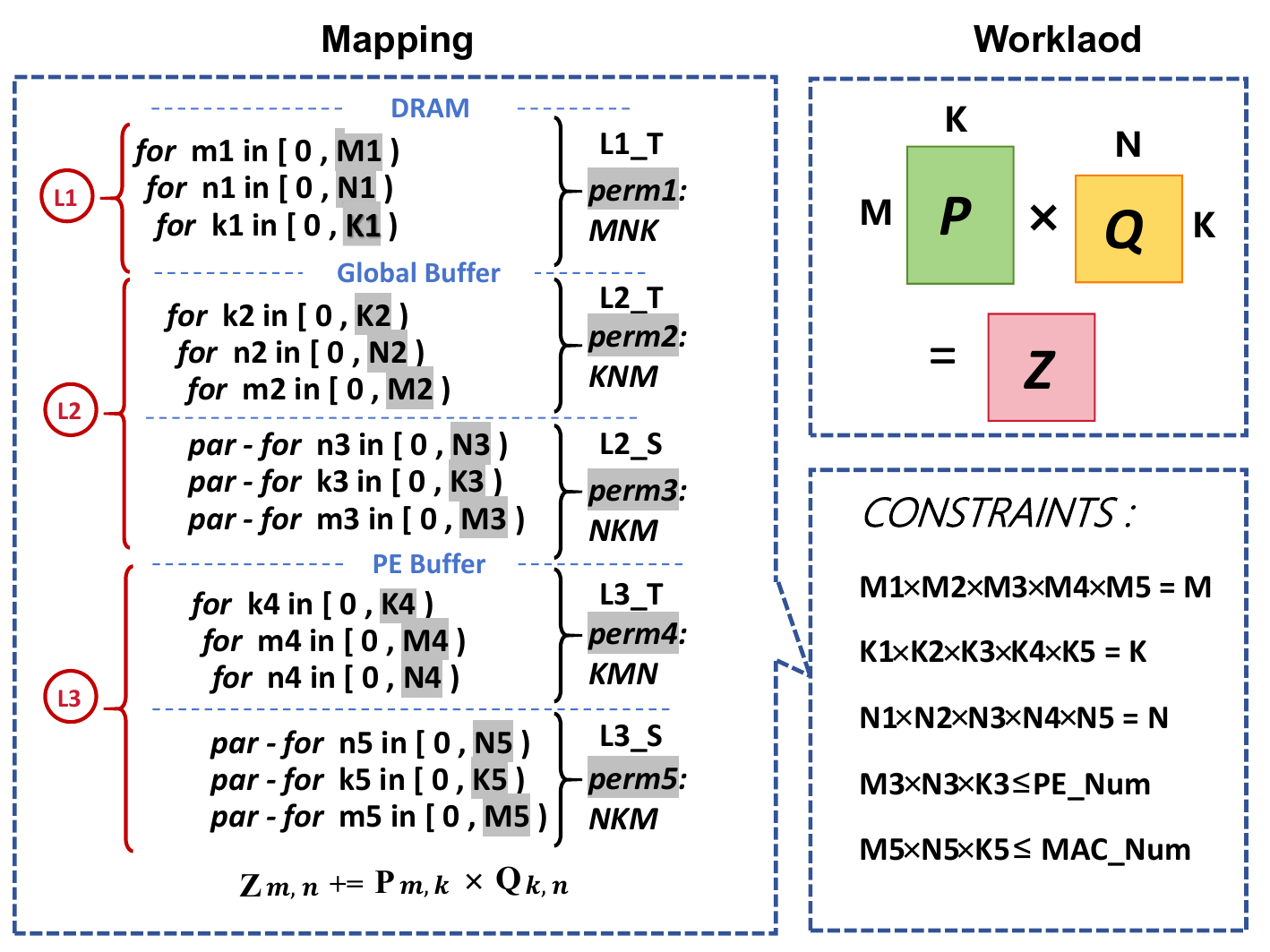}   \vspace{-3mm}
    \caption{This figure demonstrates the complete mapping of a matrix multiplication on a 3-level storage architecture. The mapping include five levels: $L1\_T$, $L2\_T$, $L2\_S$, $L3\_T$, and $L3\_S$. Each mapping level contains three loops, with their upper bounds defined by $M1$\midtilde$M5$, $K1$\midtilde$K5$, and $N1$\midtilde$N5$, respectively. The ordering of loops within each mapping level is denoted as \textit{perm1\midtilde perm5}.
}
    \label{fig_mapping_2}  \vspace{-4mm}
\end{figure}


\begin{figure*}
    \centering
    \includegraphics[width=0.7\linewidth]{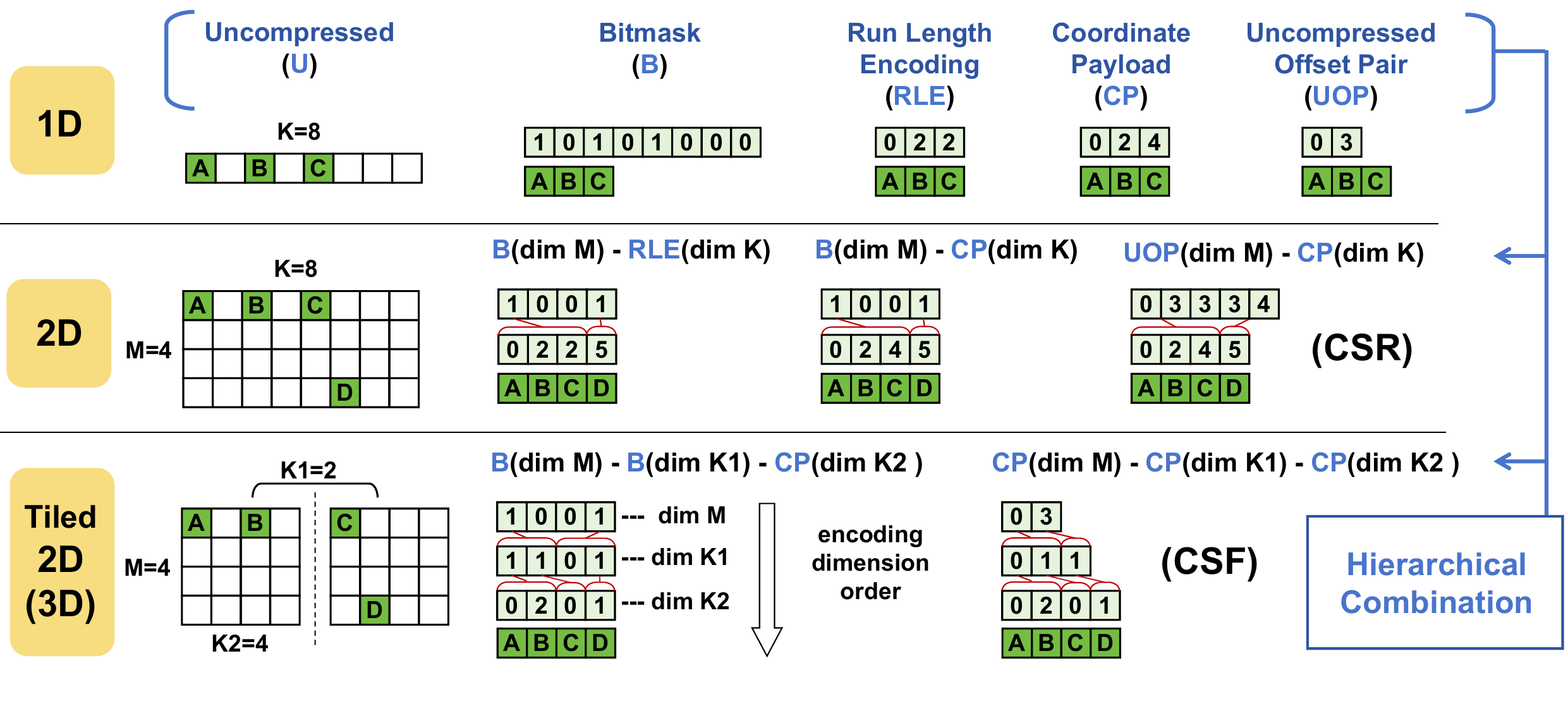} \vspace{-3mm}
    \caption{Hierarchical compression schemes for sparse data representation. The figure illustrates 1D, 2D, and tiled 2D (3D) data structures. Each level employs different encoding methods. Bitmask (B): Tracks the presence of nonzero elements using binary indicators. Run Length Encoding (RLE): Represents consecutive zeros with their lengths. Coordinate Payload (CP): Encodes the index of each nonzero element. Uncompressed Offset Pair (UOP): Stores offset pairs to quickly locate nonzero values. During the mapping process, \textit{dimension tiling} introduces additional dimensions, transforming the tensor into a higher-dimensional structure. Compression data involves two components: meta-data (light-colored sections) and data (dark-colored sections). The meta-data encodes the positions of nonzero elements, while the data section stores the values of these nonzero elements.}
    \label{fig_compress_format} \vspace{-3mm}
\end{figure*}

\subsection{Design Space}

\subsubsection{Mapping in Design Space}

Fig.~\ref{fig_mapping_2} shows the mapping on a 3-level storage architecture. When the workload involves the multiplication of matrix $P$ ($M \times K$) and matrix $Q$ ($K \times N$), the complete mapping format is illustrated on the left side of the figure. The mapping can be further divided into five mapping  levels: $L1\_T$, $L2\_T$, $L2\_S$, $L3\_T$, and $L3\_S$.  Each mapping level consists of three loops, whose upper bounds are determined by $M1$\midtilde$M5$, $K1$\midtilde$K5$, and $N1$\midtilde$N5$, respectively. These bounds are derived through \textit{dimension tiling} from the matrix dimensions $M$, $K$, and $N$. To illustrate the principle of \textit{dimension tiling}, consider a simple example: a single-loop operation with a total iteration count of 6. This loop can be restructured into two nested loops with iteration counts of 2 and 3 ($6 = 2 \times 3$). In general, the product of iteration counts for each dimension across the five mapping levels must equal the tensor size in that dimension. Mathematically, this is expressed as:
$M1 \times M2 \times M3 \times M4 \times M5 = M$ (Similarly for $K$ and $N$).
Additionally, the permutations of these loops within each mapping level are represented as $\textit{perm1} \midtilde \textit{perm5}$. These orders define the sequence of data transfer and processing across the architecture.

\subsubsection{Sparse Strategy in Design Space} 

Sparse Strategy includes compression format and S/G mechanism.


\textbf{Compression Format.} As shown in Fig.~\ref{fig_compress_format}, 
\textcolor{black}{for 1D tensors, commonly used per-dimension formats include: Bitmask (B), Run Length Encoding (RLE)\cite{rle_cite}, Coordinate Payload (CP)\cite{2021Sparseloop} and Uncompressed Offset Pair (UOP)\cite{2021Sparseloop}.} For multidimensional sparse tensors, their compression format is formed by hierarchically combining multiple 1D compression formats. For example, in the case of a 2D sparse matrix with dimensions $M$ and $K$, one can apply a format like UOP(dim M) - CP(dim K), which corresponds to the classical Compressed Sparse Row (CSR) format. \textcolor{black}{In addition, UOP needs to be used with other fomat and the number of non-zero elements in a row is represented by the difference between two elements.}


\begin{figure}
    \centering
    \includegraphics[width=1\linewidth]{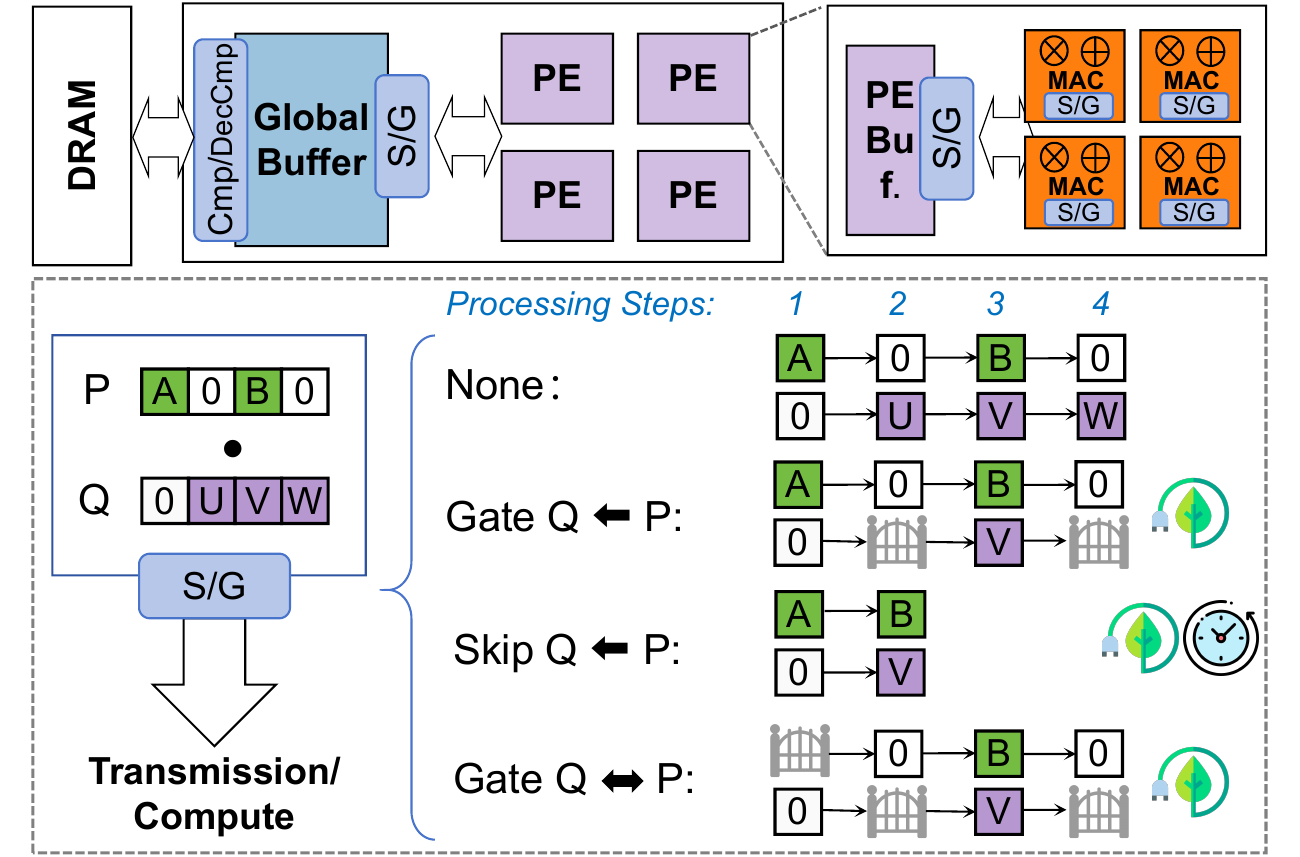}
    \caption{Illustration of the skipping/gating mechanism within a three-level storage architecture. The figure demonstrates how data streams and executed operands are processed under different strategies: None, Gate, and Skip. Gating can eliminate redundant data processing to save energy. Skipping 
 bypasses data processing involving zero elements, resulting in savings in both energy and time. Cmp/Decmp stands for compress and decompress }
    \label{fig_SG_strategy}
\end{figure}

\textbf{Skipping/Gating Mechanism.} This section elaborates that the S/G mechanism can be applied to storage and compute unit, filtering unnecessary data transmissions and computations, thereby enhancing overall performance.

As shown in Fig.~\ref{fig_SG_strategy},
\ding{182} \textit{Gating}: gating can eliminate unnecessary data processes (transmissions or computations), saving energy consumption, but can not reduce cycles. \textit{Gate P $\leftarrow$ Q} represents that a element of P is processed only when the corresponding element of Q is non-zero. 
\ding{183} \textit{Skipping}: The skipping mechanism bypasses processes of zero-element, which can save both energy and time. For instance, \textit{Skip P $\leftarrow$ Q} skips processes cycles for corresponding zero elements in Q. Since skipping requires rapidly locating the next effectual operation to jump to, it generally demands more complex hardware compared to gating. For instance, ExTensor’s intersection unit employs smart look-ahead optimizations to efficiently identify effectual operations in real time \cite{extensor}.

Additionally, a double-sided intersection mechanism, such as \textit{Skip/Gate P $\leftrightarrow$ Q}, checks metadata of both operands. If either operand is zero, neither is processed, requiring more sophisticated hardware.



\subsection{The Characteristics of Design Space}
\begin{figure}
    \centering
    \includegraphics[width=1\linewidth]{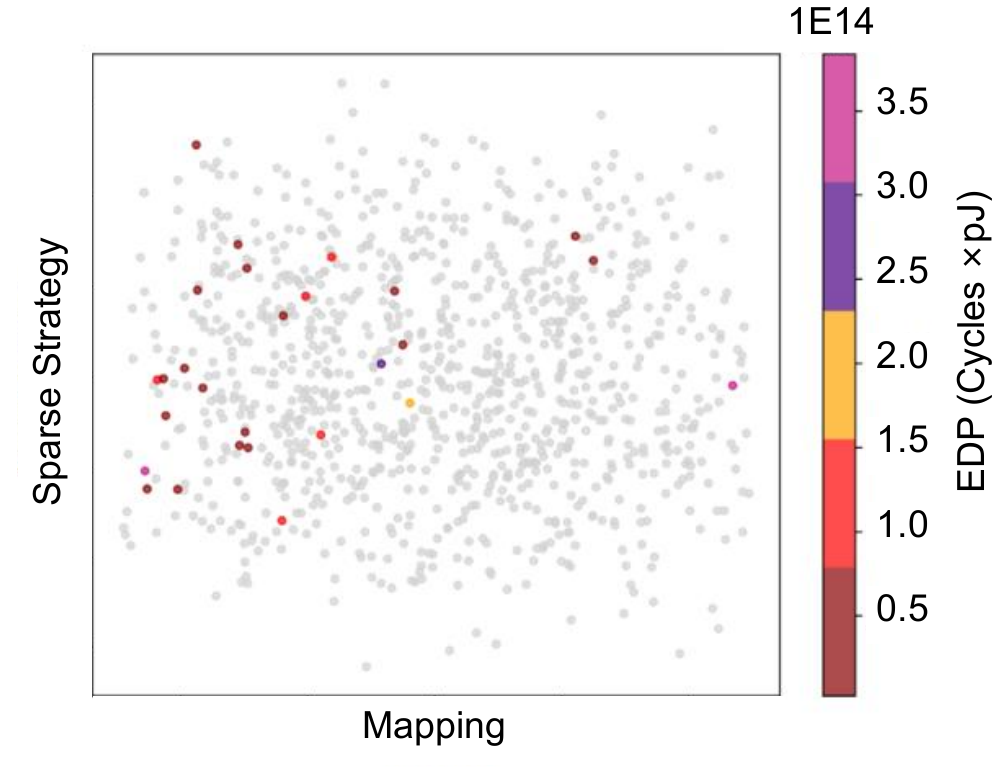}
    \caption{  
    When the workload is \textit{bibd} (a Sparse Matrix-Matrix Multiplication (SpMM) from Deepbench \cite{deepbench}), 1000 random samples are taken from the design space. We apply Principal Component Analysis (PCA) to reduce the dimensionality of the \textit{sparse strategy} variables to 1D, obtaining the vertical axis, while the \textit{mapping} variables are reduced to form the horizontal axis.
In the scatter plot, colored points represent valid design points, whereas gray points indicate invalid ones.
     The EDP for each design point is generated by evaluation model (TimeloopV2 \cite{2021Sparseloop}). 
} \vspace{-3mm}
    \label{fig_sparse_ds}
\end{figure}


All possible combinations of design points constitute the design space of sparse tensor accelerators. Both theoretical analysis and experimental results reveal the following characteristics of the design space:

\subsubsection{Vast design space.} For example, considering a three-level storage sparse tensor accelerator for the workload $P_{32 \times 64} \times Q_{64 \times 48} = Z_{32 \times 48}$, the number of potential mappings can reach $O(10^{28})=(A_{3}^{3})^5 \times 32^5 \times 64^5 \times 48^5$ (5 represents the number of mapping levels and $A_{3}^{3}$ represents the possible permutations of three dimensions,). Additionally, the possible sparse strategies for a three-level storage sparse tensor accelerator can reach $O(10^{13})=5^{15} \times 7^3$ (the calculation method is based on the encoding scheme described in the next section.). The combination of these factors results in  a design space containing $O(10^{41})$ possible accelerator designs.

\subsubsection{Enormous invalid design points.} As illustrated in Fig.~\ref{fig_sparse_ds}, the design space contains a significant number of invalid design points (represented by gray points), which overwhelmingly surround a small number of valid design points (colored points). This problem arises from two main reasons: \ding{182} many design points require hardware resources exceeding the constraints. For example, when the sparse strategy configures tensor $P$ as uncompressed, requiring more storage, and the mapping keeps $P$ stationary, the P must be stored on-chip. This can exceed the available SRAM capacity, making the design infeasible. \ding{183} The mapping and sparse strategy are incompatible. For instance, the dimension order of the compression format is inconsistent with the mapping.

\subsection{Baseline Methods}
In the field of Design Space Exploration (DSE), the following optimization methods are commonly used, and they will also serve as the baselines in this paper.
\begin{itemize}
    \item Particle Swarm Optimization (PSO): PSO is a population-based optimization method inspired by the social behavior of birds flocking or fish schooling. PSO has been used to explore optimal power-execution time tradeoff during architectural synthesis of data (and control) intensive applications \cite{pso_dse}. 

    

    \item Monte Carlo Tree Search (MCTS) is a simulation-based decision algorithm that iteratively builds a search tree while balancing exploration and exploitation. It consists of four stages: selection, expansion, simulation, and backpropagation, prioritizing high-potential branches. Prior work used it to explore mapping and hardware allocation in dense accelerator \cite{2023REMAP}.
    

    \item Proximal Policy Optimization (PPO): PPO is a reinforcement learning algorithm that balances exploration and exploitation by adjusting the policy parameters within a constrained range. PPO can be employed to explore the best  hardware resources assignment in DNN accelerator \cite{kao2020confuciux}. 
    \item Deep Q-Network (DQN): DQN is a reinforcement learning algorithm that combines Q-learning with deep neural networks to handle high-dimensional state spaces, which is also utilized for automated parameter tuning \cite{gebrekidan2024autonomous}. 
\end{itemize}


\subsection{Challenge of DSE}

The aforementioned characteristics of the design space pose significant challenges for the exploration of SpTA accelerator design. Specifically, on one hand, the enormous design space makes brute-force search algorithms infeasible. For instance, considering the design space of $O(10^{41})$ mentioned in subsection B, assuming 1000 designs are evaluated per second \cite{2021Sparseloop}, evaluating all the design points would take over $10^{30}$ years. On the other hand, the large number of invalid design points within the design space leads to low efficiency in heuristic search algorithms. For example, in MCTS, each node contains a large number of invalid branch, making it difficult for tree to guide the exploration direction \cite{2023REMAP}. To address this issue, we propose the SparseMap framework, which will be described in detail in the following section.



\section{Proposed SparseMap Framework}
Evolution Strategy (ES) is a promising optimization method characterized by strong robustness, independence from gradients and adaptability to complex high-dimensional problems \cite{es1,es2,es3,es4}. 
Previous researches have demonstrated that ES is highly competitive in the field of hardware architecture parameter search. For example, GAMMA \cite{2020GAMMA} uses ES for DNN mapping parameter search, and ConfuciuX \cite{kao2020confuciux} applies it for hardware resource allocation strategy optimization.
However, directly applying standard ES for SpTA accelerator design space exploration poses significant challenges. 

%
%
%
%
%

\subsection{Chanllenges with Evolution Strategy}



\textbf{Genetic Encoding of Large Design Space.} %
In this problem, directly using variable values as gene encoding would result in an excessively large search space, as discussed in Section III.B. Furthermore, in SpTA accelerator design, problem variables are subject to constraints. For example, the \textit{dimension tiling} of $M$ must satisfy 
$M1 \times M2 \times M3 \times M4 \times M5 = M$. Therefore, the key challenge in gene encoding is how to fully cover the valid solution space while satisfying the constraints, all while minimizing the exploration space as much as possible.

\textbf{Genetic Encoding of Hierarchical Mapping and Sparse Strategy.} %
Previous researches \cite{ga_corelative1, ga_corelative2, ga_corelative3} on ES have shown that, when encoding solutions into genomes, it is critical to ensure that similar genomes represent similar solutions. Otherwise, minor genetic mutations may lead to drastic transitions in the solution space, preventing ES from conducting local searches effectively and thereby reducing search efficiency. Notably, in this problem, both mapping and compression format exhibit hierarchical structures. For instance, compared to inner \textit{for} loops, the outer \textit{for} loops in mapping have a more significant impact on accelerator behavior. Under these conditions, adhering to the aforementioned ES encoding principle requires a carefully designed encoding scheme, which poses a considerable challenge.

\textbf{Initialization in Spaces with Numerous Invalid Points.} %
Standard Evolution Strategy commonly adopts initialization methods such as random initialization ~\cite{rand_init1,rand_init2,rand_init3}, Latin Hypercube Sampling (LHS) initialization ~\cite{cube_init1,cube_init2,cube_init3,cube_init4}, and others.  
However, these methods overlook two critical factors: \ding{182} the prevalence of invalid design points in the design space, and \ding{183} the varying degrees of impact that different design variables have on accelerator performance.  
As a result, these approaches often fail to ensure both the validity and diversity of the initial population, particularly with respect to the more influential variables.  
An initial population containing a large number of invalid individuals or lacking diversity can lead to low exploration efficiency or premature convergence to suboptimal solutions.


\textbf{Balancing Global and Local Search.} 
Typically, in ES, the early exploration phase should emphasize global search to maintain population diversity and avoid local optima, while the later phase should prioritize local search to accelerate convergence. However, in this problem, different variables have varying degrees of impact on accelerator design. Assigning the same mutation probability to all variables may lead to inefficient exploration (i.e., focusing on less influential variables in the early stage, making it difficult to ensure population diversity, and shifting to more influential variables in the later stage, hindering rapid convergence). Therefore, balancing global and local search remains a key challenge in this problem.

\textbf{Dead Offspring.} We refer to individuals that represent invalid designs as \textit{dead individuals}, characterized by a fitness value of zero. In standard ES, offspring are generated through random crossover and mutation. Due to the stochastic alteration of gene sequences, even offspring derived from highly fitted parents have a probability of becoming dead individuals. For example, if a certain dimension of a tensor is relatively dense, changing the compression format of this dimension from bitmask to coordinate payload may result in a significant increase in metadata, since coordinate payload requires more bits to encode the positions of nonzero elements. It may exceed buffer size and bandwidth limits.

This not only severely impacts population diversity but also results in identical fitness values, leading to a lack of selection directionality. Consequently, search efficiency is reduced, making it difficult to find high-quality solutions.

\begin{figure}[!t]
    \centering
    \includegraphics[width=1.0\linewidth]{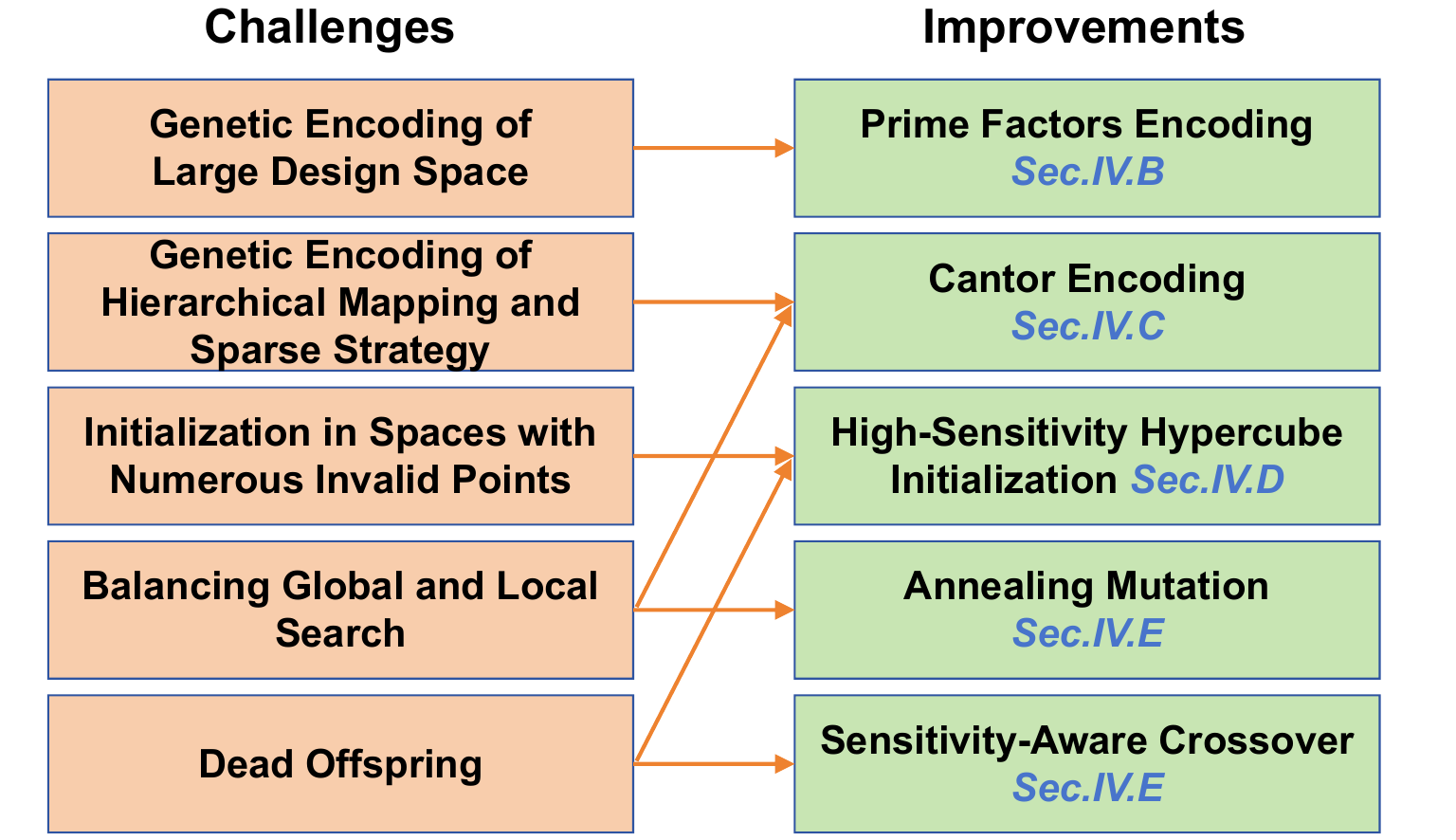}
    \caption{The challenges and corresponding improvements.}
    \label{fig:c_i} \vspace{-3mm}
\end{figure}

To address the aforementioned challenges, we propose a series of corresponding improvements, as shown in Fig.~\ref{fig:c_i}, which will be detailed in Subsections B\midtilde E.




 \subsection{Improvement: Prime Factors Encoding.}  

\begin{figure}[t]
    \centering
    \includegraphics[width=0.8\linewidth]{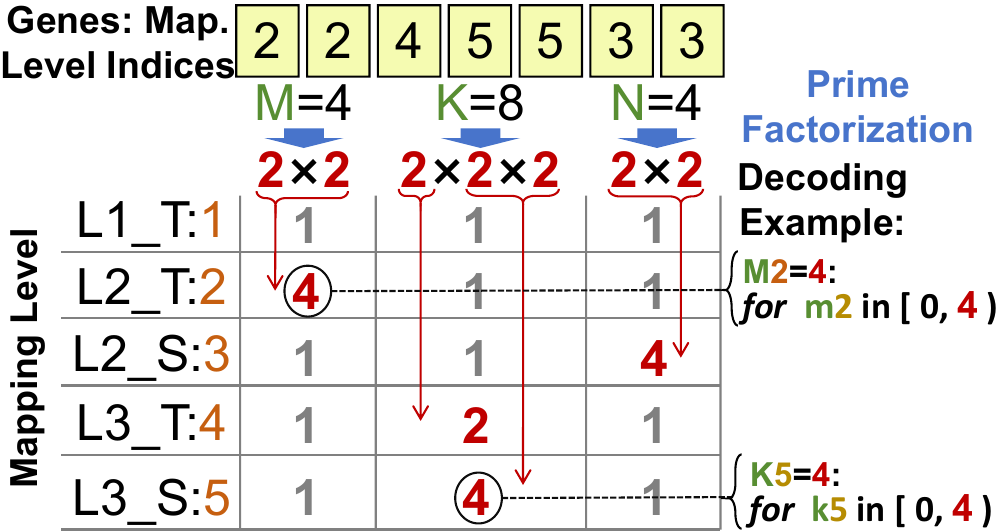}
    \caption{Illustration of prime factors encoding for \textit{dimension tiling}.}
    \label{fig:factor_encoding} \vspace{-3mm}
\end{figure}

To fully cover valid solutions while minimizing the exploration space under \textit{dimension tiling} constraints, we apply prime factors encoding. Specifically, we decompose the dimensions involved in SpTA into a sequence of prime factors, aligning them one-by-one to form a prime factor sequence. We use a genes segment to represent the mapping level (1-5) to which each prime factor is assigned.  The five mapping levels correspond to $L1\_T$, $L2\_T$, $L2\_S$, $L3\_T$, and $L3\_S$. For example, as shown in Fig.~\ref{fig:factor_encoding}, consider the case where $M=4=2 \times 2$, $K=8=2 \times 2 \times 2$, and $N=4=2 \times 2$. The two prime factors of $M$ (both 2) are assigned to the second mapping level, leading to $M2=2 \times 2$, corresponding to $for$ m2 in [0,4) in the mapping. Among the three prime factors of $K$ (all 2), the first is assigned to the fourth mapping level ($K4=2$), while the remaining two are assigned to the fifth mapping level ($K5=2 \times 2=4$). Similarly, $N3=2 \times 2=4$. Any unmentioned $M1\midtilde M5$, $K1\midtilde K5$, and $N1\midtilde N5$ (i.e., those without assigned prime factors) default to 1. Considering that input tensors may be padded in practical scenarios, if a dimension size is a large prime number, We replace it with the nearest larger composite number to ensure it can be factorized.

Compared to directly encoding the numerical values of $M1\midtilde M5$, $K1\midtilde K5$, and $N1\midtilde N5$ using 15 genes, our method significantly reduces the exploration space. If we encode the 15 numerical values directly, where $M1\midtilde M5$ range from 1 to 4, $K1\midtilde K5$ range from 1 to 8, and $N1\midtilde N5$ range from 1 to 4, the total number of possible genes combinations would be $4^5 \times 8^5 \times 4^5$. However, among these, only 7,875 combinations satisfy the \textit{dimension tiling} constraints (e.g., $M1 \times M2 \times M3 \times M4 \times M5 = M$), accounting for merely 0.000023\% of the total search space. This would result in extensive invalid exploration when using an Evolution Strategy (ES) algorithm. In contrast, prime factors encoding ensures that every generated gene combination inherently satisfies the \textit{dimension tiling} constraints. This significantly reduces the exploration space, improving both search efficiency and convergence speed.

 \subsection{Improvement: Cantor Encoding.}

\textcolor{black}{To ensure that similar genomes represent similar \textit{for} loops order under hierarchical mapping, we employ \textit{cantor encoding} to represent the permutation of dimensions within each mapping level}. Taking the three dimensions $M$, $K$, and $N$ as an example, we use the formula \eqref{eq:cantor_encoding} to encode a permutation into a single integer in the range $1$ to $6$ (i.e., $A^3_3$), where 1 corresponds to the permutation $MKN$. The cantor encoding formula is defined as:
\begin{equation} \label{eq:cantor_encoding}
\text{Encoding} = \sum_{i=1}^{d} (a_i - 1) \times (d - i)! + 1  
\end{equation} 
where \( d \) represents the number of dimensions, and \( a_i \) denotes the rank of the \( i \)-th dimension among the remaining unused dimensions, starting from 1.   

\begin{figure}[t]
    \centering
    \includegraphics[width=0.9\linewidth]{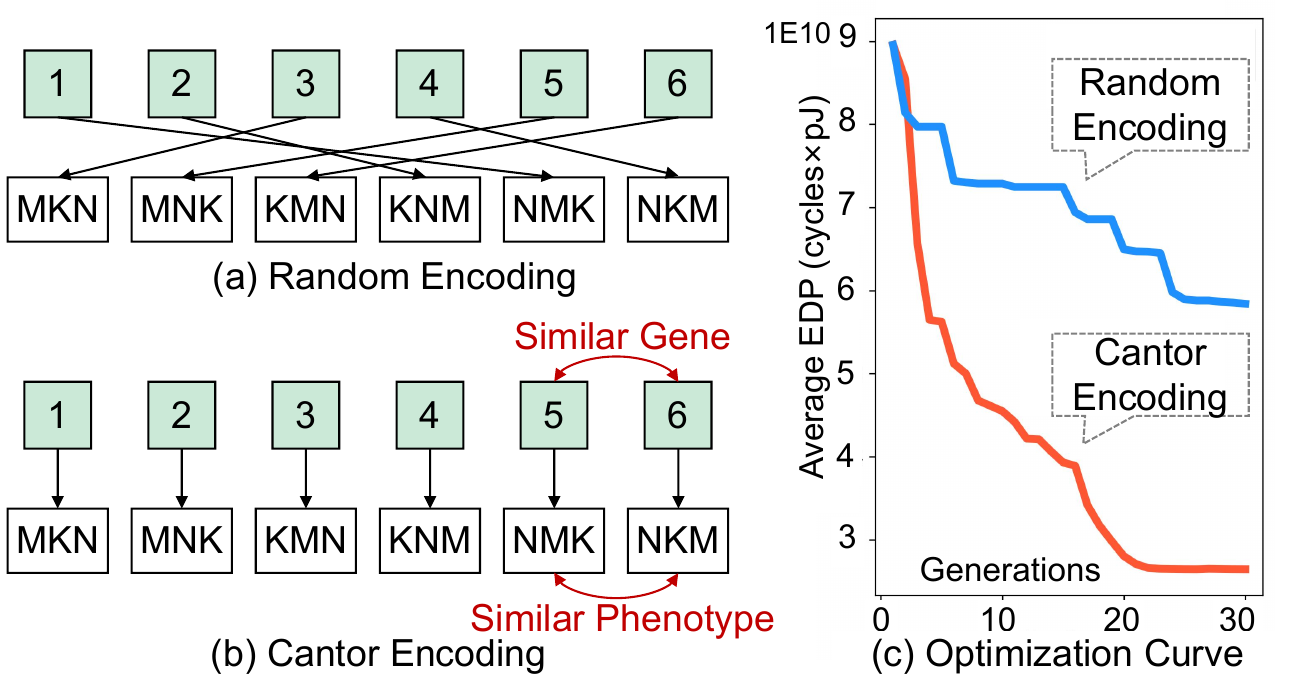}
    \caption{ (a) \textit{Cantor encoding} and (b) random encoding for permutations of mapping. 
(c) Convergence curve of different encoding. The workload is SpMM with ID: mm3. The hardware platform is cloud and the optimization objective is EDP.
}  \vspace{-3mm}
    \label{fig:cnator_encoding}
\end{figure}

Compared to random encoding, cantor encoding offers the advantage that the magnitude of gene differences reflects the actual differences in accelerator mappings. Specifically, in cantor encoding, the effect of encoding from left to right decreases. Correspondingly, in accelerator mapping, the outer loop order has a more significant influence on accelerator design. As illustrated in Fig.~\ref{fig:cnator_encoding}, (a) depicts random encoding, where two adjacent gene values (3 and 4) represent significantly different permutations, making it difficult to perform local exploration. In contrast, (b) shows cantor encoding, where genes differences correlate positively with permutation differences. The convergence curves in the right plot demonstrate that random encoding struggles to achieve rapid convergence. 

 \subsection{Improvement: High-Sensitivity Hypercube Initialization.} 

\textcolor{black}{\textbf{High-Sensitivity Gene Calibration.}}
 In the design of sparse tensor accelerators, different variables have varying impacts on accelerator performance. Some variables, when altered, can lead to an order-of-magnitude change in performance or even directly affect the validity of the design. These variables are referred to as \textit{high-sensitivity} variables (e.g., the permutation of the for-loop in the L1 mapping, which directly influences the DRAM access pattern). Variables other than the high-sensitivity ones are referred to as \textit{low-sensitivity} variables.
 
\textcolor{black}{The positions of high-sensitivity variables vary across different hardware architectures. Therefore, we adopt the Monte Carlo-based High-Sensitivity Gene Calibration to identify them. Specifically, as shown in Eq.~\eqref{eq:sensitivity_i}, for each gene \(v\), we apply the control variable method: gene $v$ is sampled via Monte Carlo\cite{montecarlo}, while all other genes except $v$ are fixed to a specific combination (which can be randomly selected), and the cost model is used to evaluate the fitness values, forming a value set denoted as $V^i$. From \(V^i\), we remove invalid design points to construct \(V_d^i\), where \(N^i\) denotes its size. Then, two values \(v_1\) and \(v_2\) are randomly selected from \(V_d^i\), and the EDP variation ratio between them is computed. The average EDP variation ratio is taken as the sensitivity of gene \(v\).}

\begin{equation}
\label{eq:sensitivity_i}
\textcolor{black}{S_i(v) = \frac{1}{N^i} \sum_{v_1,v_2 \in V_d^i} \frac{|EDP(v_1)-EDP(v_2)|}{|v_1-v_2| \cdot \min\{EDP(v_1), EDP(v_2)\}}}
\end{equation}

\textcolor{black}{To account for potential correlations between different genes, we repeat the above process \(I\) times with different random combinations of the other genes, and take the average of all \(I\) trials to obtain a robust sensitivity estimate, as shown in Eq.~\eqref{eq:sensitivity}:}

\begin{equation}
\label{eq:sensitivity}
\textcolor{black}{S(v) = \frac{1}{I} \sum_{i=1}^{I} S_i(v)}
\end{equation}
\textcolor{black}{As shown in Eq.~\eqref{eq:sens_high} and~\eqref{eq:sens_low}, given a gene sequence \(\{v_1, v_2, v_3, \ldots, v_n\}\) consisting of \(n\) genes, and the corresponding sensitivity sequence \(\{S(v_1), S(v_2), \ldots, S(v_n)\}\), genes with sensitivity values higher than a predefined threshold are identified as \textit{high-sensitivity} variables, while the remaining ones are considered \textit{low-sensitivity} variables. The threshold is defined as:
\[
\textcolor{black}{\frac{3}{4} \cdot \left( S_{\max} - S_{\min} \right) + S_{\min}}
\]
where \(S_{\max}\) and \(S_{\min}\) denote the maximum and minimum values in the sensitivity sequence, respectively. This \(3/4\) threshold is empirically derived and has been shown to effectively identify high-sensitivity variables across all tested cases.}

\begin{equation}
\label{eq:sens_high}
\textcolor{black}{V_{\text{high}} = \left\{ v \,\middle|\, S(v) > \frac{3}{4} \cdot \left( S_{\max} - S_{\min} \right) + S_{\min} \right\}}
\end{equation}

\begin{equation}
\label{eq:sens_low}
\textcolor{black}{V_{\text{low}} = \left\{ v \,\middle|\, S(v) \leq \frac{3}{4} \cdot \left( S_{\max} - S_{\min} \right) + S_{\min} \right\}}
\end{equation}

\textcolor{black}{In addition, within $V_d^i$ , we also collect valid combinations of low-sensitivity genes. These combinations are later used in the high-sensitivity hypercube initialization process to determine the value assignments for low-sensitivity genes.}

\textbf{High-Sensitivity Hypercube Initialization.} To ensure both validity and diversity in the more influential variables (we call them high-sensitivity variables) within the initial population, we propose the \textit{high-sensitivity hypercube initialization}. On one hand, we sample the initial population within a search space where high-sensitivity variables serve as the coordinate axes. Since high-sensitivity variables are typically associated with critical design parameters and significantly impact accelerator performance, ensuring diversity in these variables within the initial population allows ES to better capture the structural characteristics of the design space in the early exploration stage, guiding the evolution process toward more promising directions. On the other hand, we uniformly partition the search space into multiple hypercubes, ensuring that the initialized individuals are distributed across different hypercubes. This further enhances the diversity of high-sensitivity variable combinations in the initial population, improving the global search capability of ES and mitigating the risk of premature convergence to local optima.


 
\begin{figure}[!t]
    \centering
    \includegraphics[width=0.9\linewidth]{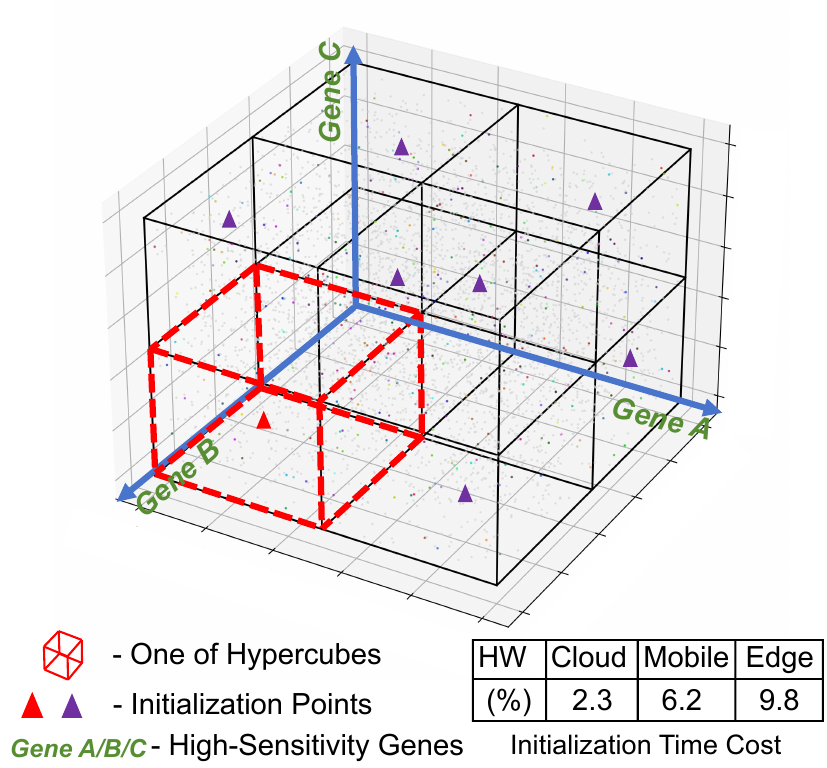}
    \caption{Illustration of \textit{high-sensitivity hypercube initialization} in our work. The design space is divided into several quadrants, with high-sensitivity segment genes (Gene A/B/C) used as coordinate axes (only three coordinate axes are shown for clarity). }
    \label{fig:inital} \vspace{-3mm}
\end{figure}

 Specifically, we uniformly divide the design space into multiple hypercubes based on high-sensitivity variables, with the number of hypercubes empirically set to 100. Within each hypercube, a small budget of random search (set to 20) is performed to obtain a valid individual for initialization. \textcolor{black}{Intuitively, a larger design space necessitates a proportionally greater random search budget and a corresponding increase in the number of hypercubes.} As shown in Fig.~\ref{fig:inital}, high-sensitivity variables Gene A, B, and C serve as spatial axes, and a valid individual (red triangle) is selected from the red hypercube, with similar operations in others. Notably, \textit{high-sensitivity hypercube initialization}'s random search incurs low overhead due to the limited budget, accounting for less than 10\% of the total search time on average.

\begin{figure}[!t]
    \centering
    \includegraphics[width=1.0\linewidth]{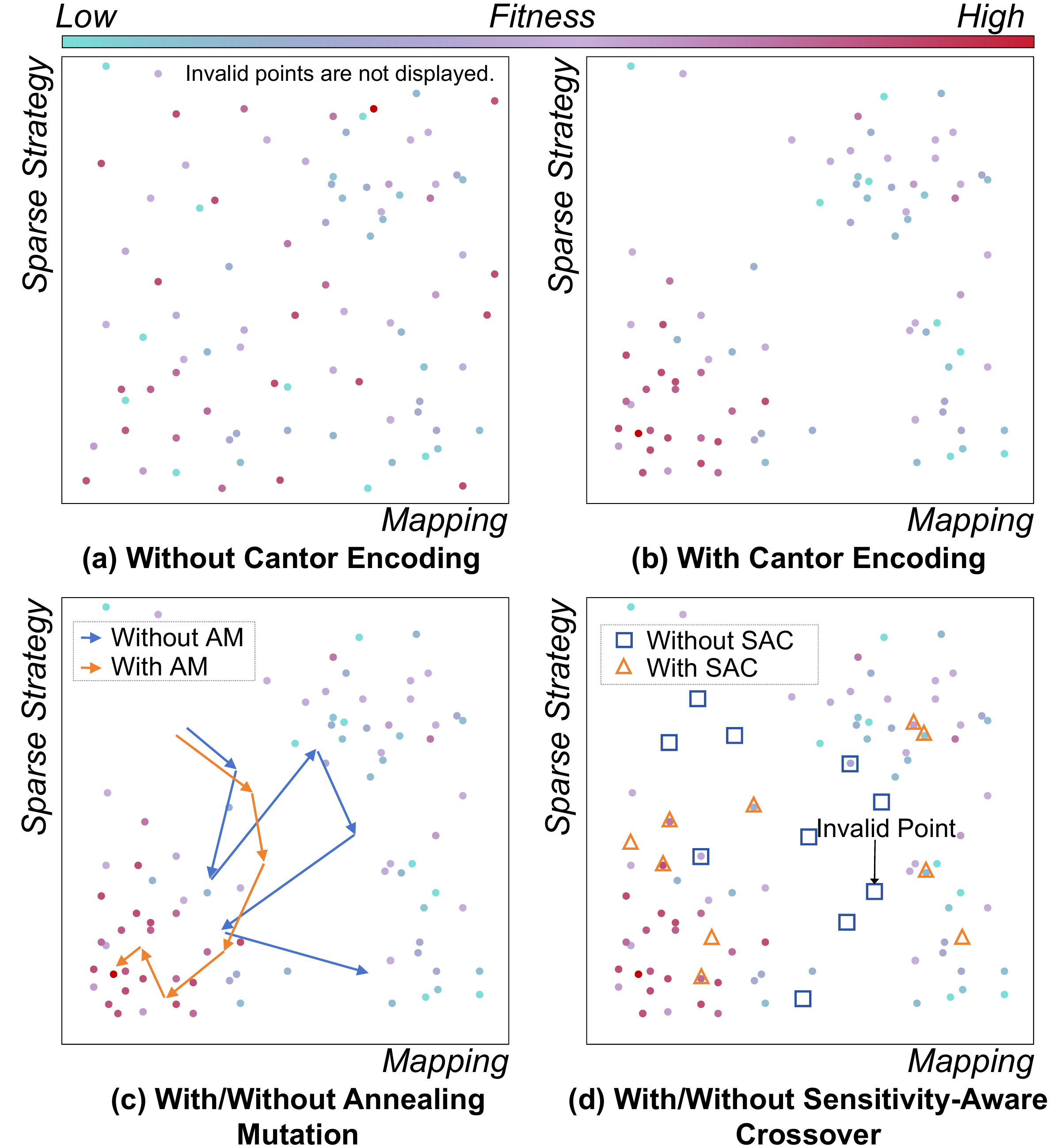}
    \caption{(a) Design space without cantor encoding, where the fitness variation between adjacent design points is large and hinder ES, posing challenges for evolution. (b) Design space with cantor encoding, where design points with similar fitness values are more clustered. (c) Evolutionary trends in the later stage of ES algorithms. Using annealing mutation is more beneficial for convergence. (d) Triangles and squares represent newly explored points in an iteration. Using SAC can reduce the probability of exploring invalid points.} \vspace{-3mm}
    \label{fig:method_eff}
\end{figure}

\subsection{Improvement: Customized Evolutionary Operators}

\textbf{Annealing Mutation.}
To balance global and local search in ES, we propose \textit{annealing mutation}. Given that different variables impact accelerator design to varying degrees, annealing mutation first classifies genes into high-sensitivity and low-sensitivity segments. During evolution, when an individual undergoes mutation, the selection of which gene segment to mutate is determined based on probabilities $P_h$ and $P_l$, which denote the probability of mutation occurring in the high-sensitivity and low-sensitivity gene segments, respectively. To adapt the search process dynamically, $P_h$ follows an annealing schedule, gradually decreasing according to:

\begin{equation}
P_h(g) = 0.8 \cdot e^{-\phi} \cdot (1-\phi),\ \phi = g/G
\end{equation}
where $g$ represents the current generation, and $G$ is the total number of generations. Conversely, $P_l$ follows the opposite trend:
\begin{equation}
P_l(g) = 1 - P_h
\end{equation}
This adaptive strategy ensures that in the early stages of evolution, the algorithm explores high-sensitivity genes more frequently, enhancing population diversity and preventing premature convergence to local optima. As the search progresses, the mutation focus shifts toward low-sensitivity genes, facilitating faster convergence. By dynamically adjusting mutation probabilities, annealing mutation effectively balances exploration and exploitation, improving the overall efficiency of the ES algorithm in optimizing accelerator design parameters.

\textbf{Sensitivity-Aware Crossover.} 
To alleviate the problem of \textit{dead offspring}, we propose \textit{sensitivity-aware crossover}. As discussed earlier, high-sensitivity genes have a significant impact on accelerator performance, meaning that large variations in these genes are more likely to produce invalid individuals. To prevent the random fragmentation of these high-sensitivity segments of the genome, we adopt the natural boundaries of high-sensitivity gene segments as crossover points. By aligning the crossover operation with these boundaries, the likelihood of generating \textit{dead offspring} is significantly reduced, thereby reducing the probability of \textit{dead offspring} occurrence. 




\begin{figure*}[!t]
    \centering
    \includegraphics[width=0.8\linewidth]{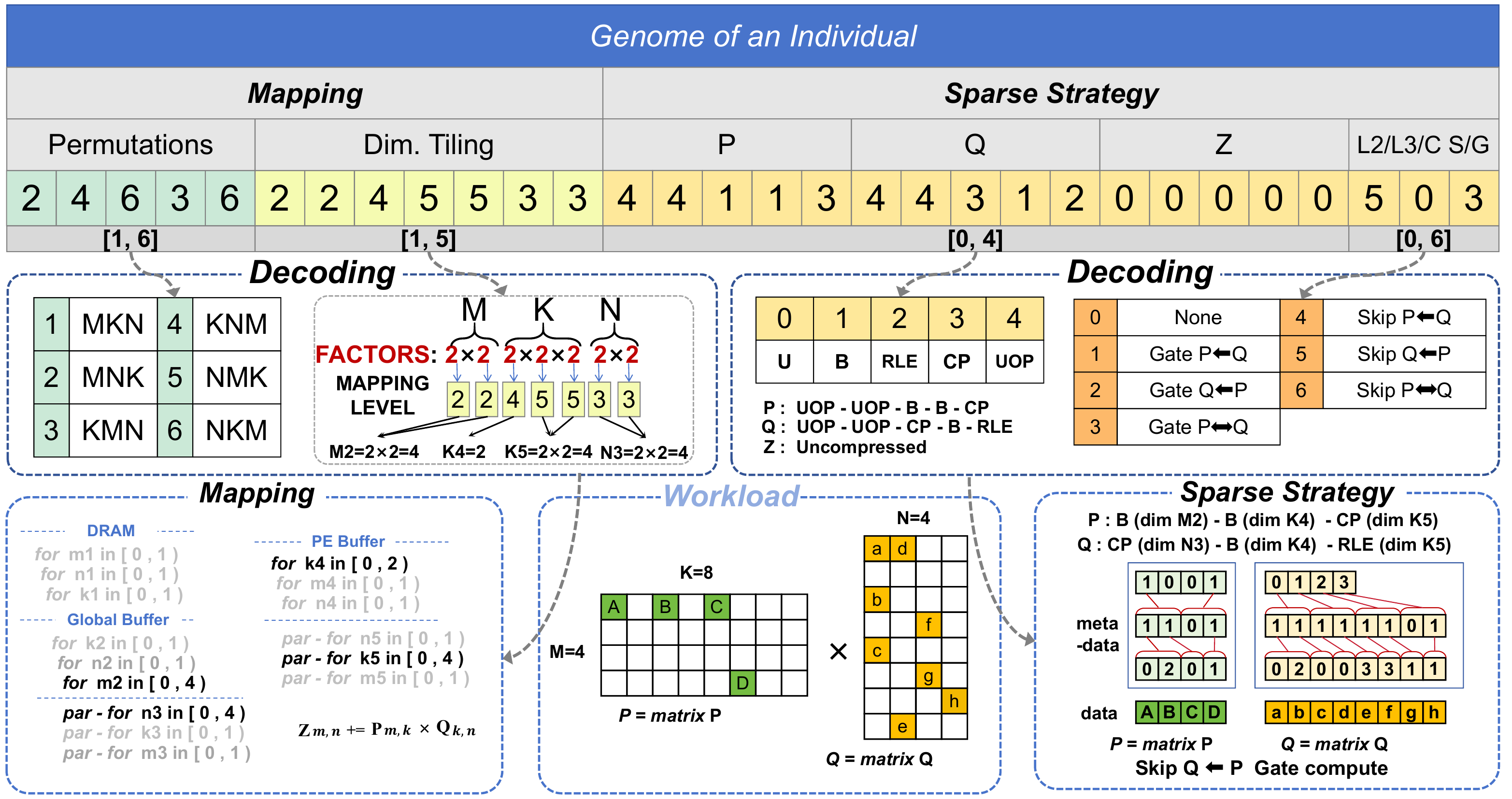}  \vspace{-3mm}
    \caption{At the top is the genome of an individual. Permutations are \textit{cantor encoding} results, indicating the loop order of dimensions across five mapping levels (perm1 to perm5 from left to right). Dim. Tiling is the result of \textit{prime factors encoding}, representing the dimension tiling of tensors. $P$, $Q$, and $Z$ denote the compression formats of the two input tensors and the output tensor. L2/L3/C-S/G represents the skipping or gating mechanism at L2, L3 and compute unit. The middle shows the correspondence between genes and design variables. The genes decodes into the mapping and sparse strategy at the bottom.} \vspace{-3mm}
    \label{fig:encoding}
\end{figure*}

\begin{figure*}[!t]
    \centering
    \includegraphics[width=0.8\linewidth]{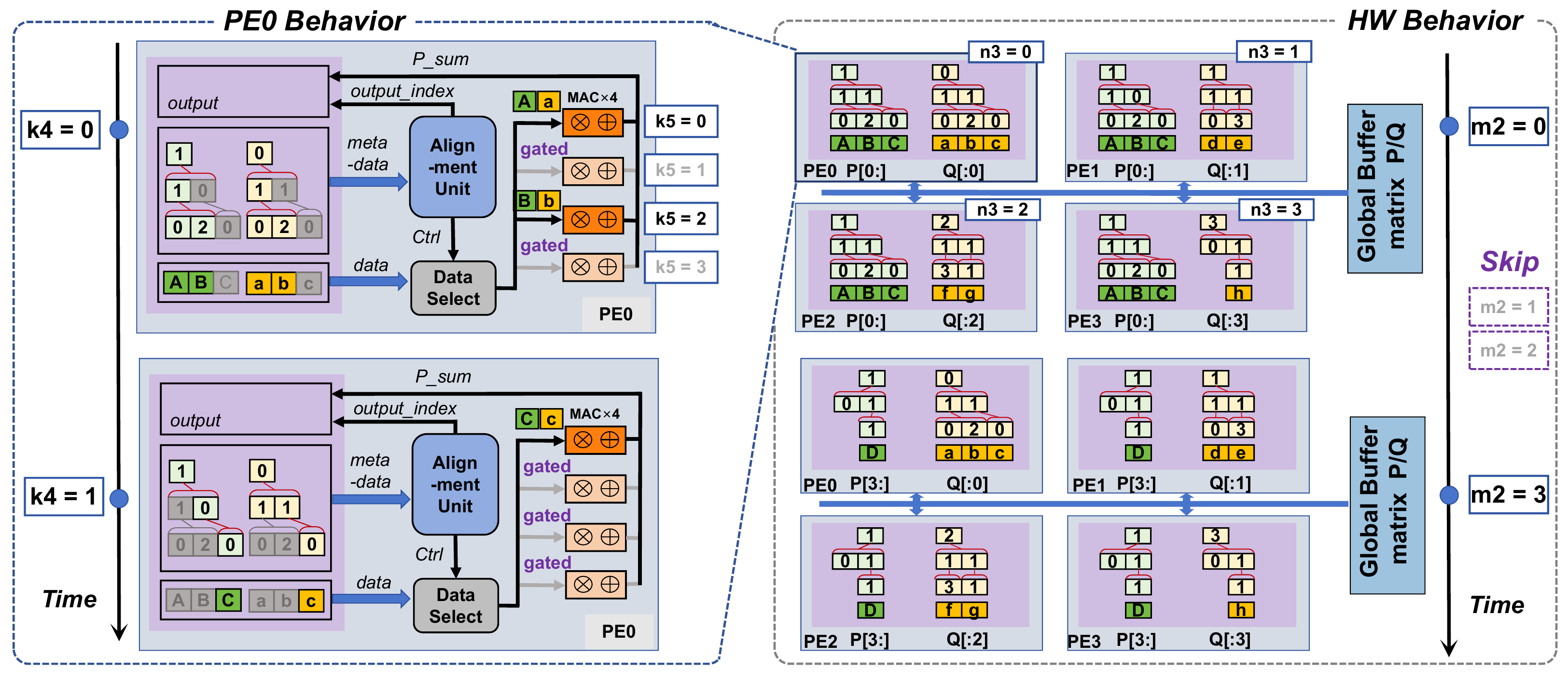} \vspace{-3mm}
    \caption{This figure presents the hardware accelerator design corresponding to the mapping and sparse strategy in Fig.~\ref{fig:encoding}. The right subfigure depicts the overall hardware behavior, while the left subfigure illustrates the behavior of PE0.} \vspace{-3mm}
    \label{fig:encoding_hw}
\end{figure*}

\subsection{Encoding-Decoding Scheme}






    

To map each sparse tensor accelerator design to a 1D genome, we developed a specialized gene encoding scheme, as shown in the top section of Fig.~\ref{fig:encoding}. 


The genome of an individual is divided into two segments: the \textit{mapping} segment and the \textit{sparse strategy} segment.

1) \textbf{Mapping segment}. As mentioned earlier, the mapping segment is composed of two sub-segments: Permutations and Dim. Tiling. The `Permutations' sub-segment encodes the loop order of tensor dimensions across the five mapping levels (perm1 to perm5 from left to right) using \textit{cantor encoding}. The `Dim. Tiling' sub-segment represents the dimension tiling of tensors, obtained through \textit{prime factors encoding}. During decoding, the loop order derived from the `Permutations' is combined with the `Dim. Tiling' to construct the complete mapping format, as illustrated in the bottom left part of Fig.~\ref{fig:encoding}.

2) \textbf{Sparse strategy segment.} The Sparse Strategy segment is further divided into five sub-segments: \textit{P}, \textit{Q}, \textit{Z}, \textit{L2/L3/C-S/G}. Specifically:

- \textit{P, Q and Z sub-segments.}
These sub-segments represent the compression formats of the two input tensors and one output tensor, respectively. Each gene in these segments has a value range of 0\midtilde4, where each value corresponds to a specific 1D compression format. The mapping between gene values and compression formats is provided in the table beneath the respective sub-segment. As described earlier, the compression format of a sparse tensor is formed by hierarchically combining multiple 1D compression formats.

For example, consider the sparse matrix $P$ illustrated in the figure, where the mapping specifies $M = M1 \times M2 \times M3 \times M4 \times M5 = 1 \times 4 \times 1 \times 1 \times 1$ and $K = K1 \times K2 \times K3 \times K4 \times K5 = 1 \times 1 \times 1 \times 2 \times 4$. This indicates that $P$ is not split along the $M$ dimension (only $M2 > 1$), but is split along the $K$ dimension into $K4 \times K5 = 2 \times 4$. Consequently, 1D compression formats need to be specified for $M2$, $K4$, and $K5$. Specifically, the compression formats are determined by the last three genes in the \textit{P} sub-segment. In this example, the formats specified for $M2$, $K4$, and $K5$ are $B(\text{dim } M2) - B(\text{dim } K4) - CP(\text{dim } K5)$.

To limit the length of an individual’s genome and simplify hardware design for encoding and decoding, the gene length of each \textit{P}, \textit{Q}, and \textit{Z} sub-segment is fixed at 5. If the compressed tensor has fewer than 5 dimensions, the encoding uses the last few genes. If the number of dimensions exceeds 5, any dimensions beyond the first 5 are automatically assigned the uncompressed format (\textit{UOP}). A gene value of 0 indicates that the tensor is uncompressed.

- \textit{L2/L3/C-S/G sub-segment}
This sub-segment encodes the Skipping/Gating (S/G) mechanism corresponding to the global buffer($L2$), PE buffer($L3$) and compute unit($C$). Gene values range from 0\midtilde6, with the mapping between values and S/G mechanism provided in the table below this sub-segment. In Fig.~\ref{fig:encoding}, this sub-segment of the genome is 5, 0 and 3, which means that \textit{Skip Q $\leftarrow$ P} is applied at the global buffer, no S/G strategy is used at the PE buffer, while \textit{Gate Q $\leftrightarrow$ P} is applied in MAC.



According to the aforementioned encoding and decoding scheme, the genome of the individual shown at the top of Fig.~\ref{fig:encoding} can be decoded into the corresponding mapping and sparse strategy, as illustrated at the bottom of the figure. Combining these two aspects yields a specific accelerator design, whose hardware behavior is depicted in Fig.~\ref{fig:encoding_hw}. Specifically, we introduce the four outer-to-inner loops in the mapping with an iteration count greater than 1 (in black) as follows: 

1) \texttt{for m2 in [0,4)} implies that four rows of matrix $P$ are broadcast in four steps (corresponding to $m2 = 0, 1, 2, 3$) to the entire PE array, including both meta-data and data. Since the second and third rows of $P$ are empty, and the S/G mechanism \textit{Skip Q $\leftarrow$ P} is applied at the global buffer, meaning that the accelerator skips $m2 = 1$ and $m2 = 2$ to save time and energy.

2) \texttt{par-for n3 in [0,4)} indicates that the meta-data and data of four columns of $Q$ are distributed across the buffers of PE0 to PE3 (corresponding to $n3 = 0, 1, 2, 3$ unfolded via spatial parallelism).

3) \texttt{for k4 in [0,2)} represents dividing the computation into two parts, processing half of the data in each iteration when processing the dot product between a row of $P$ and a column of $Q$ (corresponding to $k4 = 0$ and $k4 = 1$). Taking PE0 as an example (left side of Fig.~\ref{fig:encoding_hw}), when computing the dot product between $P[0,:]$ and $Q[:,0]$, the first four pairs of elements are processed first, followed by the latter four pairs.

4) \texttt{par-for k5 in [0,4)} denotes that the computation of four pairs of elements is distributed across four different MAC units within each PE (corresponding to $k5 = 0, 1, 2, 3$). Due to the presence of \textit{Gate Q $\leftrightarrow$ P}, the MAC unit remains idle when either element in a pair is zero. For example, in the upper left part of Fig.~\ref{fig:encoding_hw}, when $k4 = 0$, only the first and third MAC units in PE0 perform actual computations (corresponding to $k5 = 0$ and $k5 = 2$), while the remaining MAC units stay idle to save energy.

\begin{figure}[!t]
    \centering
    \includegraphics[width=1.0\linewidth]{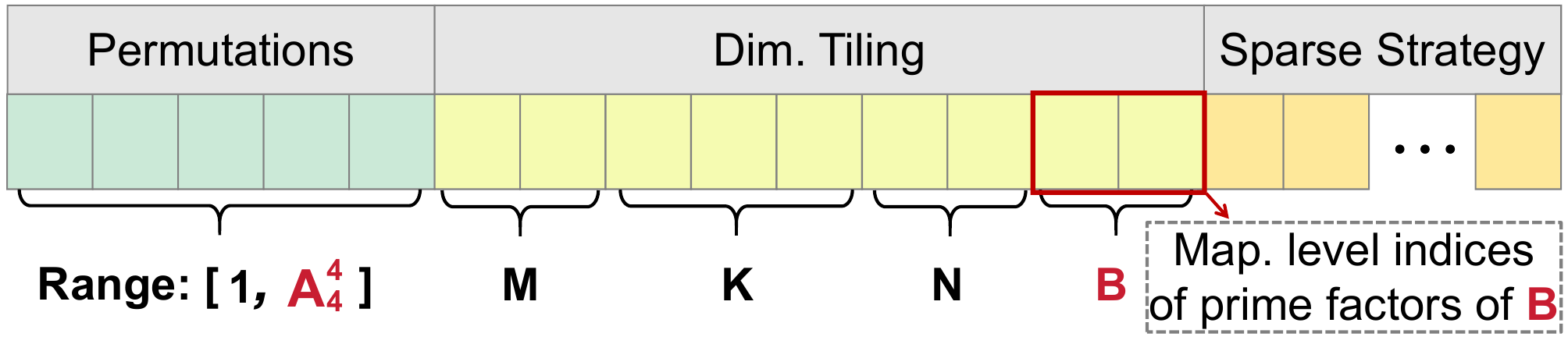}
    \caption{An example of SparseMap supporting a SpTA workload involving four dimensions, where dimensionz B is newly added.}
    \label{fig:multi_dim_en}
\end{figure}

\subsection{Support for Multi-Dimensional Workload}

In the previous description, we used SpMM involving three dimensions as example. When the SpTA workload involves more dimensions, SparseMap can still support. Specifically, SparseMap first obtains the dimension information of the tensors involved in the SpTA from the user-provided workload description file. Then, it configures the genome of corresponding length based on the dimension sizes. For instance, as shown in Fig.~\ref{fig:multi_dim_en}, by adding a batch dimension 
$B$ to the matrix multiplication, the range of genetic values for the permutation segment changes from $[1, A^3_3]$ to $[1, A^4_4]$ (the new workload involves four dimensions). The dim. tiling segment will include additional mapping level indices corresponding to the prime factors of $B$.





\begin{figure}
    \centering
    \includegraphics[width=1\linewidth]{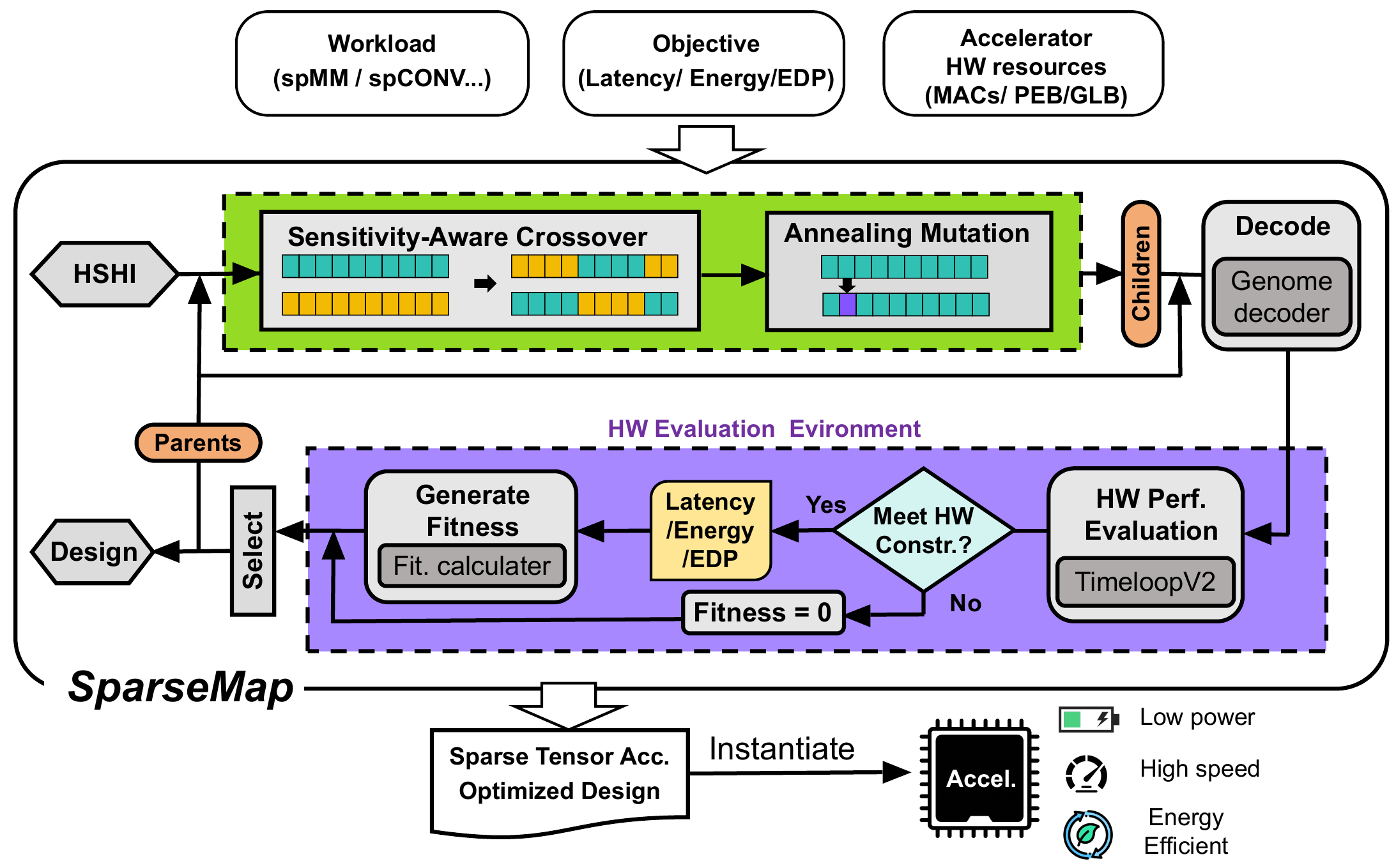}
    \caption{Overview of the SparseMap framework. The framework combines evolution strategy (\textit{High-Sensitivity Hypercube Initialization} (HSHI) \textit{sensitivity-aware crossover}, \textit{annealing mutation}, and operations) with a hardware evaluation environment. The output design of SparseMap can be instantiated into a sparse tensor accelerator, ensuring low power, high speed, and energy efficiency.}
    \label{Sparseloop_framework}
\end{figure}

\subsection{Flow for Evolution Strategy}

As shown in Fig.~\ref{Sparseloop_framework}, the evolutionary strategy in SparseMap iteratively optimizes to seek the efficient SpTA accelerator design through the following process:

\textbf{Initialization:} We initialize the population using the \textit{high-sensitivity hypercube initialization}  described in Subsection D.

\textbf{Crossover:} We select high-fitness individuals from the population as the parent set. In crossover, pairs of individuals are randomly selected from the parent set, and the sensitivity-aware crossover described earlier is performed to generate new offspring, which are then added to the population.


\textbf{Mutation:} Each new offspring has a certain probability of undergoing mutation, following the \textit{annealing mutation} method proposed in this paper.


\textbf{Evaluation and Selection:} After crossover and mutation, we evaluate the population by interacting with the HW evaluation environment, which will be described in the next subsection. The HW evaluation environment provides feedback on the fitness of each individual. We select the population eligible to enter the next generation based on their fitness ranking.

\subsection{SparseMap Framework.} 



As shown in Fig.~\ref{Sparseloop_framework}, SparseMap receives input files from the user, including workload optimization objectives, and hardware resource constraints. Based on the feedback from the hardware evaluation environment, after a specified number of iterations (defined by the user) in the ES flow, it ultimately produces the optimized sparse tensor accelerator design (including mapping and sparse strategy) as output, which can be instantiated to actual accelerator by engineers to achieve the best performance. 


\textbf{Workload.} Sparse tensor workload is defined by input files, which includes operator types, tensor dimensions, sparsity, zero elements distribution, etc.

\textbf{Optimization Objective.} SparseMap's aim is to minimize a user-defined hardware performance metric, which can include energy, delay or energy-delay product (EDP).

\textbf{Constraints.} The automated search in SparseMap is conducted under hardware resource constraints. Specifically, these constraints include the number of MACs, the capacity of each buffer level, components and process used in the accelerator and so on.


\textbf{Evaluation Environment.} The hardware evaluation environment in SparseMap is built on TimeloopV2 (aka. Sparseloop)\cite{2021Sparseloop}, an evaluation model for sparse tensor accelerators. TimeloopV2 achieves an average error of 0.1\% to 8\% across representative accelerator designs and workloads. 

\section{Experiments}

To validate the optimization capability of the SparseMap framework, we conducted comparative experiments with baseline optimization methods and two advanced works in the same domain (SAGE \cite{ocf3_extending} and Sparseloop Mapper \cite{2021Sparseloop}). To validate the effectiveness of the proposed genetic encoding, custom evolutionary operators and initialization method, we conducted ablation studies. The baseline for the ablation experiments is the standard ES (with latin hypercube sampling initialization).

To the best of the authors' knowledge, no prior work has simultaneously explored both sparse strategies and mappings for sparse tensor accelerators. SAGE and Sparseloop Mapper are advanced works in the field of sparse tensor accelerator design space exploration. SAGE, proposed by Qin \cite{ocf3_extending}, explores the compression format of sparse tensors under the assumption that the mapping is fixed. \textcolor{black}{Sparseloop Mapper conducts mapping exploration under a manually specified sparse strategy, with mapping candidates generated in consideration of \textit{dimension tiling} constraints.}
We replicated SAGE in the evaluation environment used in this paper, calling it SAGE-like, and incorporated the manual settings of Sparseloop Mapper into its random sampling space for a more reasonable comparison. We conducted comparative experiments on the three specified hardware platforms with the workloads from Table \ref{tab_workload}, given the same search budget. Similar to previous design space exploration works \cite{2020GAMMA,2023REMAP,kao2020confuciux}, \textcolor{black}{we set the optimization goal was set as EDP (measured in cycles $\times$ pJ). The total search budget as 20,000 samples based on engineering experience. It is important to note that this total search budget is not universally optimal. As the design space scales up, the search budget must also increase dynamically to ensure convergence of the search results.}

\subsection{Hardware Platform}
In order to make the experiment more representative, we set three hardware resource constraints as shown in Table~\ref{tab_platforms}, including edge, mobile, and cloud. The hardware resources of edge and cloud are at the same level as the Eyeriss chip \cite{2016Eyeriss} and cloud TPU \cite{tpu}, respectively, while the hardware resources of mobile fall between the two. In the experimental setup, we used the same 12nm process technology as DSTC\cite{dstc}.

\begin{table}[t]
\caption{Hardware resource constrains of different platforms}
\centering
\resizebox{0.5\textwidth}{!}{ 
\begin{tabular}{|l|c|c|c|c|c|}
\hline
\textbf{Platform} & \textbf{Number of PE} & \textbf{MACs in PE} & \textbf{PE buffer} & \textbf{Global Buffer} & \textbf{DRAM Bandwidth} \\
\hline
Edge & $16\times 16$ & 1 & 1 KB & 128 KB & 16 MB/s \\
\hline
Mobile & $16\times 16$ & 64 & 32 KB & 16 MB & 32 GB/s \\
\hline
Cloud & $32\times 32$ & 64 & 128 KB & 64 MB & 128 GB/s \\
\hline
\end{tabular}
}

\label{tab_platforms}

\end{table}


\subsection{Workload}
As shown in Table \ref{tab_workload}, our workload includes two types, SpConv and SpMM, representing sparse tensor algebra. The SpConv workload comes from the convolutional layers of a VGG16 network with 50\% global pruning applied to its parameters (the fine-tuned model's accuracy is 93.22\%). The SpMM workload is partially derived from Deepbench\cite{deepbench}, representing SpMM workloads in deep learning, and partially from sparseGPT\cite{frantar2023sparsegpt}, which includes sparse MHA and MLP, representing the demand for SpMM in Large Language Models (LLM).

\begin{table*}[t]
\caption{Sparse Tensor Algebra Workloads}
\centering
\label{tab_workload}
\resizebox{\textwidth}{!}{
\begin{tabular}{|c|c|c|c|c|c||c|c|c|c|c|c|}
\hline
\textbf{ID} & \textbf{operation} & \textbf{Operater1} & \textbf{density\%} & \textbf{Operater2} & \textbf{density\%} & 
\textbf{ID} & \textbf{operation} & \textbf{Operater1} & \textbf{density\%} & \textbf{Operater2} & \textbf{density\%} \\
\hline
mm1 & SpMM & $124 \times 124$ & 78.5 & $124 \times 124$ & 78.5 &
\textcolor{black}{mm15} & \textcolor{black}{SpMM} & \textcolor{black}{$1K \times 16K$} & \textcolor{black}{60.0} & \textcolor{black}{$16K \times 16K$} & \textcolor{black}{78.0} \\
\hline
mm2 & SpMM & $171 \times 92K$ & 20.9 & $92K \times 171$ & 20.9 &
conv1 & SpConv & $3 \times 32 \times 32$ & 100 & $64 \times 3 \times 3 \times 3$ & 54.6 \\
\hline
mm3 & SpMM & $730 \times 730$ & 11.8 & $730 \times 730$ & 11.8 &
conv2 & SpConv & $64 \times 32 \times 32$ & 45.0 & $256 \times 64 \times 1 \times 1$ & 25.2 \\
\hline
mm4 & SpMM & $7.7K \times 2.6K$ & 5.0 & $2.6K \times 7.7K$ & 5.0 &
conv3 & SpConv & $128 \times 16 \times 16$ & 39.6 & $512 \times 128 \times 1 \times 1$ & 36.6 \\
\hline
mm5 & SpMM & $9K \times 9K$ & 4.1 & $9K \times 9K$ & 4.1 &
conv4 & SpConv & $128 \times 16 \times 16$ & 47.7 & $128 \times 128 \times 3 \times 3$ & 64.7 \\
\hline
mm6 & SpMM & $2.6K \times 2.6K$ & 1.1 & $2.6K \times 2.6K$ & 1.1 &
conv5 & SpConv & $1024 \times 8 \times 8$ & 40.2 & $256 \times 1024 \times 1 \times 1$ & 50.1 \\
\hline
mm7 & SpMM & $1.6K \times 4.6K$ & 0.3 & $4.6K \times 1.6K$ & 0.3 &
conv6 & SpConv & $256 \times 8 \times 8$ & 43.0 & $256 \times 256 \times 3 \times 3$ & 61.7 \\
\hline
mm8 & SpMM & $2K \times 12.3K$ & 100 & $12.3K \times 128$ & 50.0 &
conv7 & SpConv & $512 \times 4 \times 4$ & 59.0 & $2048 \times 512 \times 1 \times 1$ & 11.8 \\
\hline
mm9 & SpMM & $2K \times 12.3K$ & 100 & $12.3K \times49.2K$ & 50.0 &
\textcolor{black}{conv8} & \textcolor{black}{SpConv} & \textcolor{black}{$128 \times 64 \times 64$} & \textcolor{black}{40.0} & \textcolor{black}{$512 \times 128 \times 4 \times 4$} & \textcolor{black}{30.0} \\
\hline
mm10 & SpMM & $2K \times 49.2K$ & 100 & $49.2K \times12.3K$ & 50.0 &
\textcolor{black}{conv9} & \textcolor{black}{SpConv} & \textcolor{black}{$128 \times 64 \times 64$} & \textcolor{black}{100} & \textcolor{black}{$64 \times 128 \times 1 \times 1$} & \textcolor{black}{20.0} \\
\hline
\textcolor{black}{mm11} & \textcolor{black}{SpMM} & \textcolor{black}{$128 \times 1024$} & \textcolor{black}{0.6} & \textcolor{black}{$1024 \times 128$} & \textcolor{black}{0.6} &
\textcolor{black}{conv10} & \textcolor{black}{SpConv} & \textcolor{black}{$256 \times 64 \times 64$} & \textcolor{black}{40.0} & \textcolor{black}{$512 \times 256 \times 1 \times 1$} & \textcolor{black}{25.0} \\
\hline
\textcolor{black}{mm12} & \textcolor{black}{SpMM} & \textcolor{black}{$768 \times 64$} & \textcolor{black}{5.9} & \textcolor{black}{$64 \times 768$} & \textcolor{black}{5.9} &
\textcolor{black}{conv11} & \textcolor{black}{SpConv} & \textcolor{black}{$4 \times 32 \times 32$} & \textcolor{black}{34.0} & \textcolor{black}{$64 \times 4 \times 3 \times 3$} & \textcolor{black}{14.6} \\
\hline
\textcolor{black}{mm13} & \textcolor{black}{SpMM} & \textcolor{black}{$12.3K \times 24.6K$} & \textcolor{black}{1.0} & \textcolor{black}{$24.6K \times 12.3K$} & \textcolor{black}{1.0} &
\textcolor{black}{conv12} & \textcolor{black}{SpConv} & \textcolor{black}{$1024 \times 4 \times 4$} & \textcolor{black}{79.0} & \textcolor{black}{$64 \times 1024 \times 1 \times 1$} & \textcolor{black}{11.8} \\
\hline
\textcolor{black}{mm14} & \textcolor{black}{SpMM} & \textcolor{black}{$256 \times 512$} & \textcolor{black}{32.8} & \textcolor{black}{$512 \times 2048$} & \textcolor{black}{71.8} &
\textcolor{black}{conv13} & \textcolor{black}{SpConv} & \textcolor{black}{$256 \times 16 \times 16$} & \textcolor{black}{90.2} & \textcolor{black}{$128 \times 256 \times 1 \times 1$} & \textcolor{black}{5.1} \\
\hline
\end{tabular}}
\end{table*}

\begin{figure}[t]
    \centering
    \includegraphics[width=0.9\linewidth]{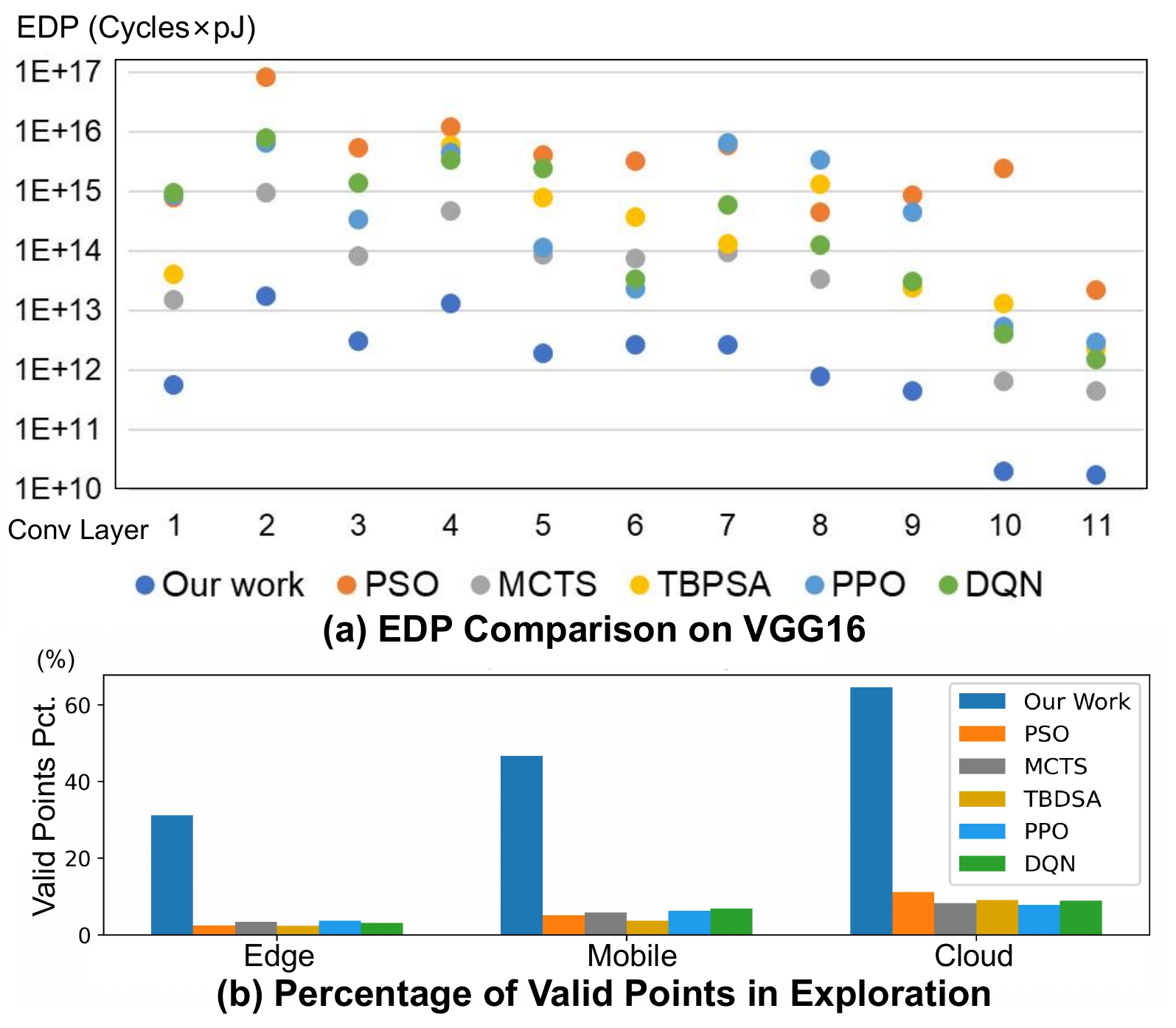}
    \caption{ (a) Comparison of the search results between our work and various baseline methods on the pruned VGG16 model under a search budget of 20,000 iterations on the Cloud platform. (b) The percentage of valid points among all explored points during the optimization process on different hardware platforms (Edge, Mobile, and Cloud), averaged across all pruned convolutional layers in VGG16.}
    \label{fig:exp_with_baseline}
\end{figure}

\subsection{Comparison with Baseline Optimization Method}

As shown in Fig.~\ref{fig:exp_with_baseline} (a), it presents a comparison of the EDP across different convolutional layers of the VGG16 model, evaluated under a search budget of 20,000 samples on the Cloud platform. The results clearly demonstrate that our proposed method, SparseMap, consistently achieves lower EDP compared to the five baseline optimization methods: PSO, MCTS, TBPSA, PPO and DQN. 


On average, SparseMap achieves a remarkable reduction in EDP by 2\midtilde5 orders of magnitude compared to PSO, MCTS, TBPSA, PPO, and DQN. The figure illustrates the consistent superiority of SparseMap across all convolutional layers, highlighting its ability to efficiently explore the design space and optimize the mapping and sparse strategy variables simultaneously.


The poor optimization performance of the baseline algorithms is mainly due to the vast search space filled with a large number of invalid design points. Consequently, a substantial portion of the search budget is consumed by evaluating these invalid solutions, significantly reducing the likelihood of discovering high-quality design points. In contrast, SparseMap employs a unique encoding scheme and customized evolutionary operators, which increase the probability of exploring valid points. Moreover, the feedback from a greater number of valid points can more effectively guide the direction of the next round of optimization. Fig.~\ref{fig:exp_with_baseline}(b) supports this hypothesis. Under the same search budget, SparseMap demonstrates a more comprehensive search over valid points compared to the baseline algorithms. This advantage is primarily attributed to SparseMap’s prime factors encoding, which greatly reduces the search space. Besides, SparseMap adopts techniques such as \textit{annealing mutation}, \textit{high-sensitivity hypercube initialization} and \textit{sensitive-aware crossover}, which collectively help maintain population diversity while mitigating the adverse effects of invalid points.

\subsection{Comparison with SAGE-like} 
 As shown in Table~\ref{tab_com}, \textcolor{black}{SparseMap achieves an average EDP reduction of $26.8\times$ on the Edge platform, $19.2\times$ on the Mobile platform, and $171.4\times$ on the Cloud platform.} This improvement is likely attributed to the limitations of SAGE-like, which focuses solely on exploring compression formats while restricting the mapping. For example, when the SpTA workload changes, such as an originally sparse input tensor becoming denser, SAGE-like approaches fail to re-explore the accelerator mapping for the new workload. As a result, the original mapping leads to frequent read/write to the tensor, increasing both latency and energy consumption. In contrast, SparseMap jointly optimizes sparse strategies and mappings, fully accounting for their mutual influence when giving a workload, leading to more balanced and efficient designs. 
%

\subsection{Comparison with Sparseloop} 
As shown in Table~\ref{tab_com}, \textcolor{black}{SparseMap outperforms Sparseloop Mapper with an average EDP reduction of $8.8\times$ on the edge platform, $4.5\times$ on the mobile platform, and $158.9\times$ on the cloud platform.} This significant improvement can be credited to the improved evolution strategy employed by SparseMap, which leverages the fitness of the previous generation's population to guide the evolution process more effectively. As a result, SparseMap is more likely to approach the global optimum compared to the random sampling-based search used by Sparseloop Mapper.

\begin{table*}[t]
\centering
\caption{Performance comparison across workloads and platforms. The optimal results are highlighted in bold. }
\label{tab_com}
\resizebox{\textwidth}{!}{
\begin{tabularx}{\textwidth}{|c|*{9}{X|}}
\hline
\multirow{2}{*}{Workload} & \multicolumn{3}{c|}{Edge} & \multicolumn{3}{c|}{Mobile} & \multicolumn{3}{c|}{Cloud} \\
\cline{2-10}
 & Sparseloop & SAGE-like & Our Work & Sparseloop & SAGE-like & Our Work & Sparseloop & SAGE-like & Our work \\
\hline
mm1 & \textcolor{black}{1.98E+10} & 2.37E+10 & \textbf{1.92E+10} & \textcolor{black}{1.17E+09} & 2.94E+09 & \textbf{3.10E+08} & \textcolor{black}{3.58E+08} & 3.34E+08 & \textbf{1.40E+08} \\
\hline
mm2 & \textcolor{black}{2.44E+14} & 4.62E+14 & \textbf{2.31E+14} & \textcolor{black}{8.84E+12} & 5.73E+13 & \textbf{6.23E+12} & \textcolor{black}{1.79E+12} & 3.64E+13 & \textbf{1.68E+12} \\
\hline
mm3 & \textcolor{black}{5.27E+12} & 1.66E+14 & \textbf{1.70E+12} & \textcolor{black}{9.39E+10} & 2.03E+12 & \textbf{4.79E+10} & \textcolor{black}{6.44E+10} & 1.05E+12 & \textbf{2.67E+10} \\
\hline
mm4 & \textcolor{black}{1.57E+16} & 5.80E+17 & \textbf{1.55E+16} & \textcolor{black}{7.73E+15} & 4.86E+15 & \textbf{2.92E+15} & \textcolor{black}{1.94E+14} & 1.54E+16 & \textbf{1.13E+14} \\
\hline
mm5 & \textcolor{black}{1.12E+18} & 1.72E+19 & \textbf{3.70E+17} & \textcolor{black}{1.17E+16} & 1.77E+16 & \textbf{8.31E+15} & \textcolor{black}{2.75E+15} & 7.91E+17 & \textbf{1.34E+15} \\
\hline
mm6 & \textcolor{black}{8.56E+12} & 4.54E+13 & \textbf{1.92E+12} & \textcolor{black}{8.24E+12} & 8.89E+12 & \textbf{2.77E+12} & \textcolor{black}{9.38E+12} & 8.86E+12 & \textbf{2.03E+10} \\
\hline
mm7 & \textcolor{black}{5.85E+11} & 6.39E+12 & \textbf{2.03E+11} & \textcolor{black}{2.21E+10} & 6.31E+10 & \textbf{9.92E+09} & \textcolor{black}{6.33E+09} & 5.34E+12 & \textbf{3.13E+09} \\
\hline
mm8 & \textcolor{black}{3.23E+16} & 1.17E+17 & \textbf{3.14E+16} & \textcolor{black}{8.26E+14} & 2.34E+15 & \textbf{4.93E+14} & \textcolor{black}{1.43E+14} & 5.43E+14 & \textbf{1.10E+14} \\
\hline
mm9 & \textcolor{black}{4.38E+21} & 9.92E+21 & \textbf{4.33E+21} & \textcolor{black}{1.06E+20} & 1.52E+20 & \textbf{7.57E+19} & \textcolor{black}{1.87E+19} & 3.05E+19 & \textbf{9.23E+18} \\
\hline
mm10 & \textcolor{black}{4.14E+21} & 8.37E+21 & \textbf{4.11E+21} & \textcolor{black}{1.56E+20} & 4.02E+20 & \textbf{1.18E+20} & \textcolor{black}{3.02E+19} & 3.65E+19 & \textbf{9.22E+18} \\
\hline
\textcolor{black}{mm11} & \textcolor{black}{3.29E+06} & \textcolor{black}{2.08E+06} & \textcolor{black}{\textbf{9.18E+05}} & \textcolor{black}{2.69E+06} & \textcolor{black}{1.07E+06} & \textcolor{black}{\textbf{6.41E+05}} & \textcolor{black}{1.11E+06} & \textcolor{black}{2.07E+08} & \textcolor{black}{\textbf{3.55E+05}} \\
\hline
\textcolor{black}{mm12} & \textcolor{black}{1.65E+10} & \textcolor{black}{4.23E+11} & \textcolor{black}{\textbf{3.16E+09}} & \textcolor{black}{7.99E+08} & \textcolor{black}{9.12E+09} & \textcolor{black}{\textbf{1.18E+08}} & \textcolor{black}{6.12E+08} & \textcolor{black}{8.65E+09} & \textcolor{black}{\textbf{3.28E+07}} \\
\hline
\textcolor{black}{mm13} & \textcolor{black}{3.37E+17} & \textcolor{black}{1.00E+18} & \textcolor{black}{\textbf{3.94E+16}} & \textcolor{black}{2.83E+16} & \textcolor{black}{3.18E+17} & \textcolor{black}{\textbf{7.26E+15}} & \textcolor{black}{1.49E+18} & \textcolor{black}{2.67E+17} & \textcolor{black}{\textbf{4.03E+14}} \\
\hline
\textcolor{black}{mm14} & \textcolor{black}{2.27E+14} & \textcolor{black}{2.31E+14} & \textcolor{black}{\textbf{5.38E+13}} & \textcolor{black}{3.28E+12} & \textcolor{black}{1.16E+13} & \textcolor{black}{\textbf{9.22E+11}} & \textcolor{black}{1.30E+12} & \textcolor{black}{5.73E+12} & \textcolor{black}{\textbf{3.77E+11}} \\
\hline
\textcolor{black}{mm15} & \textcolor{black}{3.95E+19} & \textcolor{black}{3.78E+20} & \textcolor{black}{\textbf{9.22E+18}} & \textcolor{black}{2.30E+19} & \textcolor{black}{2.03E+19} & \textcolor{black}{\textbf{2.40E+18}} & \textcolor{black}{5.92E+18} & \textcolor{black}{1.05E+19} & \textcolor{black}{\textbf{1.18E+18}} \\
\hline
conv1 & \textcolor{black}{1.15E+10} & 2.85E+10 & \textbf{3.10E+09} & \textcolor{black}{1.19E+09} & 5.88E+08 & \textbf{3.22E+08} & \textcolor{black}{3.28E+08} & 5.93E+08 & \textbf{1.01E+08} \\
\hline
conv2 & \textcolor{black}{1.23E+11} & 1.41E+11 & \textbf{1.10E+10} & \textcolor{black}{4.86E+09} & 7.57E+09 & \textbf{3.98E+08} & \textcolor{black}{5.87E+09} & 2.71E+10 & \textbf{2.19E+09} \\
\hline
conv3 & \textcolor{black}{1.22E+11} & 5.54E+11 & \textbf{1.36E+10} & \textcolor{black}{2.87E+09} & 4.87E+09 & \textbf{7.05E+08} & \textcolor{black}{8.89E+08} & 1.54E+09 & \textbf{6.60E+08} \\
\hline
conv4 & \textcolor{black}{1.68E+12} & 1.78E+12 & \textbf{2.80E+10} & \textcolor{black}{2.48E+10} & 1.12E+11 & \textbf{4.22E+09} & \textcolor{black}{2.52E+10} & 7.99E+10 & \textbf{1.63E+10} \\
\hline
conv5 & \textcolor{black}{4.58E+10} & 7.78E+11 & \textbf{1.91E+10} & \textcolor{black}{2.31E+09} & 9.51E+09 & \textbf{1.03E+09} & \textcolor{black}{2.21E+09} & 8.64E+09 & \textbf{1.16E+09} \\
\hline
conv6 & \textcolor{black}{1.37E+12} & 2.73E+12 & \textbf{9.58E+10} & \textcolor{black}{4.61E+10} & 1.74E+11 & \textbf{9.38E+09} & \textcolor{black}{3.38E+10} & 1.14E+11 & \textbf{5.75E+09} \\
\hline
conv7 & \textcolor{black}{4.58E+12} & 7.20E+12 & \textbf{2.39E+11} & \textcolor{black}{2.57E+11} & 7.68E+11 & \textbf{1.65E+10} & \textcolor{black}{2.35E+11} & 1.28E+11 & \textbf{1.33E+10} \\
\hline
\textcolor{black}{conv8} & \textcolor{black}{9.71E+15} & \textcolor{black}{2.31E+16} & \textcolor{black}{\textbf{4.14E+15}} & \textcolor{black}{2.93E+14} & \textcolor{black}{5.51E+14} & \textcolor{black}{\textbf{5.17E+13}} & \textcolor{black}{7.29E+14} & \textcolor{black}{2.75E+14} & \textcolor{black}{\textbf{1.26E+13}} \\
\hline
\textcolor{black}{conv9} & \textcolor{black}{6.20E+12} & \textcolor{black}{4.04E+12} & \textcolor{black}{\textbf{7.85E+11}} & \textcolor{black}{3.91E+10} & \textcolor{black}{8.59E+10} & \textcolor{black}{\textbf{8.54E+09}} & \textcolor{black}{1.52E+10} & \textcolor{black}{7.44E+10} & \textcolor{black}{\textbf{4.33E+09}} \\
\hline
\textcolor{black}{conv10} & \textcolor{black}{2.75E+14} & \textcolor{black}{2.20E+14} & \textcolor{black}{\textbf{4.50E+13}} & \textcolor{black}{1.02E+12} & \textcolor{black}{9.97E+12} & \textcolor{black}{\textbf{5.22E+11}} & \textcolor{black}{8.20E+11} & \textcolor{black}{3.39E+12} & \textcolor{black}{\textbf{1.37E+11}} \\
\hline
\textcolor{black}{conv11} & \textcolor{black}{2.30E+10} & \textcolor{black}{1.31E+10} & \textcolor{black}{\textbf{7.14E+08}} & \textcolor{black}{5.16E+07} & \textcolor{black}{2.04E+08} & \textcolor{black}{\textbf{7.71E+06}} & \textcolor{black}{4.56E+08} & \textcolor{black}{1.17E+08} & \textcolor{black}{\textbf{4.19E+06}} \\
\hline
\textcolor{black}{conv12} & \textcolor{black}{3.37E+09} & \textcolor{black}{2.90E+09} & \textcolor{black}{\textbf{2.33E+08}} & \textcolor{black}{2.67E+07} & \textcolor{black}{3.88E+08} & \textcolor{black}{\textbf{5.70E+06}} & \textcolor{black}{1.45E+07} & \textcolor{black}{6.31E+07} & \textcolor{black}{\textbf{9.74E+05}} \\
\hline
\textcolor{black}{conv13} & \textcolor{black}{1.43E+11} & \textcolor{black}{1.85E+11} & \textcolor{black}{\textbf{8.33E+09}} & \textcolor{black}{4.13E+08} & \textcolor{black}{3.68E+09} & \textcolor{black}{\textbf{6.83E+07}} & \textcolor{black}{1.75E+08} & \textcolor{black}{3.17E+09} & \textcolor{black}{\textbf{3.27E+07}} \\
\hline

\end{tabularx}
}
\end{table*}

\begin{figure}[t]
    \centering
    \includegraphics[width=1.0\linewidth]{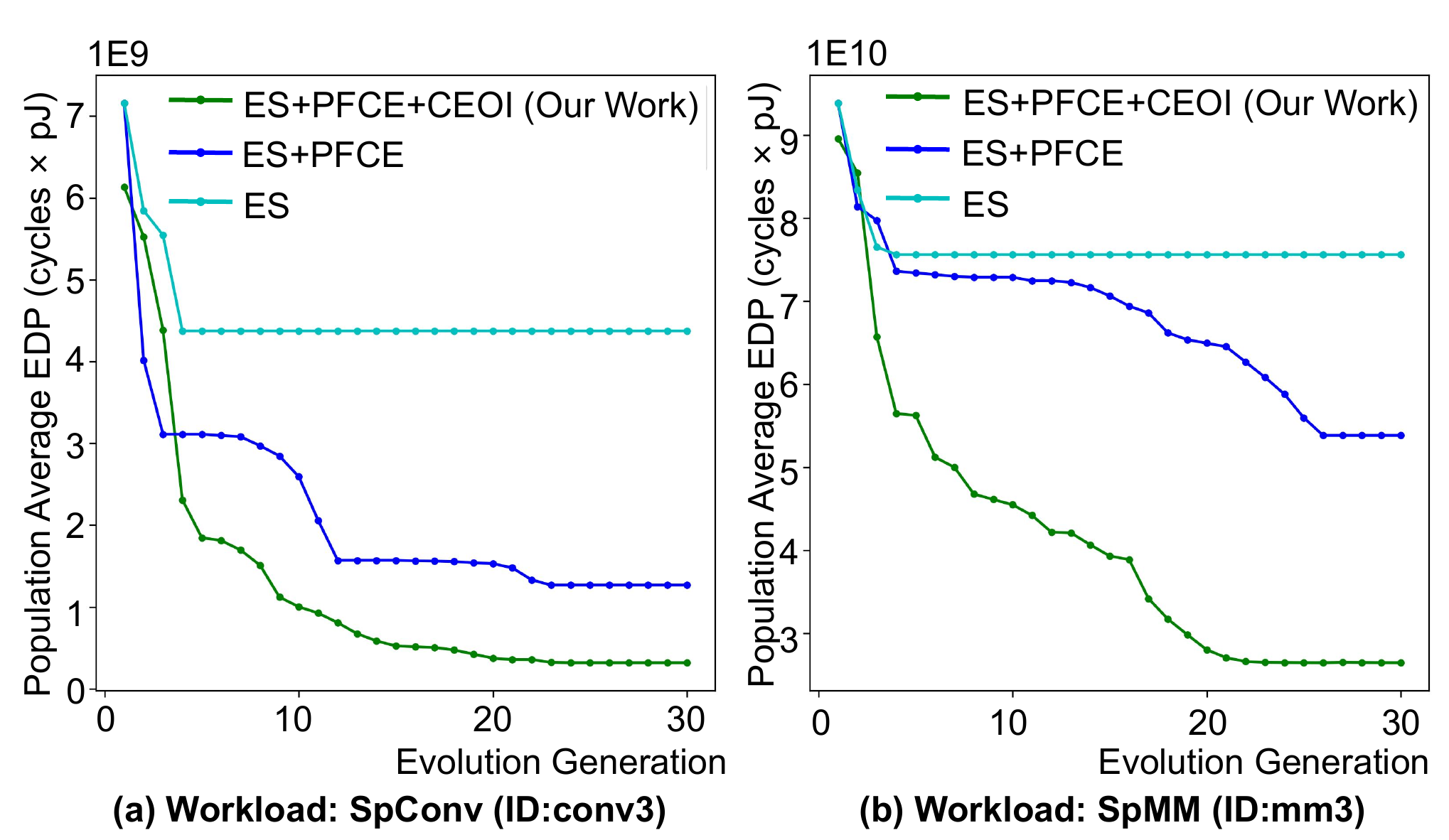}
    \caption{Convergence curves of different evolution strategies under different workloads. The hardware platform is the Cloud, and the optimization objective is EDP. ES is Evolution strategy using Latin Hypercube Sampling. PFCE is Prime Factors and Cantor Encoding. CEOI is Customized Evolutionary Operators and High-Sensitivity Hypercube Initialization.}
    \label{fig:curve}
\end{figure}

\subsection{Ablation Experiment}
As shown in Fig.~\ref{fig:curve}, we take one SpConv and one SpMM as examples, and explore the optimization of EDP using three evolution strategies under the hardware resource constraints of the cloud platform. The three evolution strategy are: 1) The standard ES with latin hypercube initialization. 2) The ES that only uses \textit{prime factors} and \textit{cantor encoding}. 3) SparseMap, which uses both customized Evolutionary Operators and \textit{high-sensitivity hypercube initialization}. The curves showing the average EDP of the population with respect to generations for the three algorithms are shown in Fig.~\ref{fig:curve}.

From the experimental results, it can be observed that compared to the standard evolution strategy, \textit{prime factors} and \textit{cantor encoding} helps avoid the introduction of dead individuals, allowing the population evolving more effectively. Additionally, the customized evolutionary operators and \textit{high-sensitivity hypercube initialization}, due to a more diverse population and balance between global and local search, gives SparseMap a better potential to find optimal designs.

\section{Conclusion}

Designing efficient sparse tensor accelerators (SpTAs) is critical for modern compute-intensive applications, but optimizing mappings and sparse strategy simultaneously under hardware constraints poses significant challenges due to the combinatorial explosion. This paper presents SparseMap, a novel Evolution Strategy-based optimization framework. With new SpTA workloads emerging at an unprecedented pace, SparseMap enables researchers to effectively explore the hardware efficiency without resorting to labor-intensive human-in-the-loop optimization processes.



\bibliographystyle{IEEEtran}

\begin{thebibliography}{10}
\providecommand{\url}[1]{#1}
\csname url@samestyle\endcsname
\providecommand{\newblock}{\relax}
\providecommand{\bibinfo}[2]{#2}
\providecommand{\BIBentrySTDinterwordspacing}{\spaceskip=0pt\relax}
\providecommand{\BIBentryALTinterwordstretchfactor}{4}
\providecommand{\BIBentryALTinterwordspacing}{\spaceskip=\fontdimen2\font plus
\BIBentryALTinterwordstretchfactor\fontdimen3\font minus \fontdimen4\font\relax}
\providecommand{\BIBforeignlanguage}[2]{{%
\expandafter\ifx\csname l@#1\endcsname\relax
\typeout{** WARNING: IEEEtran.bst: No hyphenation pattern has been}%
\typeout{** loaded for the language `#1'. Using the pattern for}%
\typeout{** the default language instead.}%
\else
\language=\csname l@#1\endcsname
\fi
#2}}
\providecommand{\BIBdecl}{\relax}
\BIBdecl

\bibitem{sparse_net_1}
M.~Abadi, P.~Barham, J.~Chen, Z.~Chen, A.~Davis, J.~Dean, M.~Devin, S.~Ghemawat, G.~Irving, M.~Isard \emph{et~al.}, ``Tensorflow: a system for large-scale machine learning,'' in \emph{12th USENIX symposium on operating systems design and implementation (OSDI 16)}, 2016, pp. 265--283.

\bibitem{sparse_net_2}
A.~Paszke, S.~Gross, F.~Massa, A.~Lerer, J.~Bradbury, G.~Chanan, T.~Killeen, Z.~Lin, N.~Gimelshein, L.~Antiga \emph{et~al.}, ``Pytorch: An imperative style, high-performance deep learning library,'' \emph{Advances in neural information processing systems}, vol.~32, 2019.

\bibitem{recommend_1}
H.~Ko, S.~Lee, Y.~Park, and A.~Choi, ``A survey of recommendation systems: recommendation models, techniques, and application fields,'' \emph{Electronics}, vol.~11, no.~1, p. 141, 2022.

\bibitem{recommend_2}
L.~Yu, J.~Huang, G.~Zhou, C.~Liu, and Z.-K. Zhang, ``Tiirec: A tensor approach for tag-driven item recommendation with sparse user generated content,'' \emph{Information Sciences}, vol. 411, pp. 122--135, 2017.

\bibitem{sci_simu}
M.~Zhao, R.~V. Panda, S.~S. Sapatnekar, T.~Edwards, R.~Chaudhry, and D.~Blaauw, ``Hierarchical analysis of power distribution networks,'' in \emph{Proceedings of the 37th Annual Design Automation Conference}, 2000, pp. 150--155.

\bibitem{gpgpu1}
K.~Zhong, Z.~Zhu, G.~Dai, H.~Wang, X.~Yang, H.~Zhang, J.~Si, Q.~Mao, S.~Zeng, K.~Hong \emph{et~al.}, ``Feasta: A flexible and efficient accelerator for sparse tensor algebra in machine learning,'' in \emph{Proceedings of the 29th ACM International Conference on Architectural Support for Programming Languages and Operating Systems, Volume 3}, 2024, pp. 349--366.

\bibitem{2016Eyeriss}
Y.~H. Chen, T.~Krishna, J.~S. Emer, and V.~Sze, ``Eyeriss: An energy-efficient reconfigurable accelerator for deep convolutional neural networks,'' \emph{Solid-state Circuits Conference}, 2016.

\bibitem{cnvlutin}
J.~Albericio, P.~Judd, T.~Hetherington, T.~Aamodt, N.~E. Jerger, and A.~Moshovos, ``Cnvlutin: Ineffectual-neuron-free deep neural network computing,'' \emph{ACM SIGARCH Computer Architecture News}, vol.~44, no.~3, pp. 1--13, 2016.

\bibitem{eyerissV2}
Y.-H. Chen, T.-J. Yang, J.~Emer, and V.~Sze, ``Eyeriss v2: A flexible accelerator for emerging deep neural networks on mobile devices,'' \emph{IEEE Journal on Emerging and Selected Topics in Circuits and Systems}, vol.~9, no.~2, pp. 292--308, 2019.

\bibitem{extensor}
K.~Hegde, H.~Asghari-Moghaddam, M.~Pellauer, N.~Crago, A.~Jaleel, E.~Solomonik, J.~Emer, and C.~W. Fletcher, ``Extensor: An accelerator for sparse tensor algebra,'' in \emph{Proceedings of the 52nd Annual IEEE/ACM International Symposium on Microarchitecture}, 2019, pp. 319--333.

\bibitem{matraptor}
N.~Srivastava, H.~Jin, J.~Liu, D.~Albonesi, and Z.~Zhang, ``Matraptor: A sparse-sparse matrix multiplication accelerator based on row-wise product,'' in \emph{2020 53rd Annual IEEE/ACM International Symposium on Microarchitecture (MICRO)}.\hskip 1em plus 0.5em minus 0.4em\relax IEEE, 2020, pp. 766--780.

\bibitem{gospa}
C.~Deng, Y.~Sui, S.~Liao, X.~Qian, and B.~Yuan, ``Gospa: An energy-efficient high-performance globally optimized sparse convolutional neural network accelerator,'' in \emph{2021 ACM/IEEE 48th Annual International Symposium on Computer Architecture (ISCA)}.\hskip 1em plus 0.5em minus 0.4em\relax IEEE, 2021, pp. 1110--1123.

\bibitem{outerspace}
S.~Pal, J.~Beaumont, D.-H. Park, A.~Amarnath, S.~Feng, C.~Chakrabarti, H.-S. Kim, D.~Blaauw, T.~Mudge, and R.~Dreslinski, ``Outerspace: An outer product based sparse matrix multiplication accelerator,'' in \emph{2018 IEEE International Symposium on High Performance Computer Architecture (HPCA)}.\hskip 1em plus 0.5em minus 0.4em\relax IEEE, 2018, pp. 724--736.

\bibitem{scnn}
A.~Parashar, M.~Rhu, A.~Mukkara, A.~Puglielli, R.~Venkatesan, B.~Khailany, J.~Emer, S.~W. Keckler, and W.~J. Dally, ``Scnn: An accelerator for compressed-sparse convolutional neural networks,'' \emph{ACM SIGARCH computer architecture news}, vol.~45, no.~2, pp. 27--40, 2017.

\bibitem{sigma}
E.~Qin, A.~Samajdar, H.~Kwon, V.~Nadella, S.~Srinivasan, D.~Das, B.~Kaul, and T.~Krishna, ``Sigma: A sparse and irregular gemm accelerator with flexible interconnects for dnn training,'' in \emph{2020 IEEE International Symposium on High Performance Computer Architecture (HPCA)}.\hskip 1em plus 0.5em minus 0.4em\relax IEEE, 2020, pp. 58--70.

\bibitem{sparten}
A.~Gondimalla, N.~Chesnut, M.~Thottethodi, and T.~Vijaykumar, ``Sparten: A sparse tensor accelerator for convolutional neural networks,'' in \emph{Proceedings of the 52nd Annual IEEE/ACM International Symposium on Microarchitecture}, 2019, pp. 151--165.

\bibitem{sparsity_recon}
J.-W. Jang, S.~Lee, D.~Kim, H.~Park, A.~S. Ardestani, Y.~Choi, C.~Kim, Y.~Kim, H.~Yu, H.~Abdel-Aziz \emph{et~al.}, ``Sparsity-aware and re-configurable npu architecture for samsung flagship mobile soc,'' in \emph{2021 ACM/IEEE 48th Annual International Symposium on Computer Architecture (ISCA)}.\hskip 1em plus 0.5em minus 0.4em\relax IEEE, 2021, pp. 15--28.

\bibitem{2021Sparseloop}
Y.~N. Wu, P.~A. Tsai, A.~Parashar, V.~Sze, and J.~S. Emer, ``Sparseloop: An analytical, energy-focused design space exploration methodology for sparse tensor accelerators,'' in \emph{2021 IEEE International Symposium on Performance Analysis of Systems and Software (ISPASS)}, 2021.

\bibitem{nvdla}
\BIBentryALTinterwordspacing
{NVIDIA}, ``{NVDLA Deep Learning Accelerator},'' NVIDIA Corporation, Tech. Rep., 2020, white Paper. [Online]. Available: \url{http://nvdla.org}
\BIBentrySTDinterwordspacing

\bibitem{stc_nvidiaA100}
\BIBentryALTinterwordspacing
NVIDIA, ``Nvidia a100 tensor core gpu architecture,'' NVIDIA Corporation, Tech. Rep., 2020, white Paper. [Online]. Available: \url{https://resources.nvidia.com/en-us-tensor-core/nvidia-ampere-architecture-whitepaper}
\BIBentrySTDinterwordspacing

\bibitem{wos}
P.~Liu and Y.~Wang, ``A low-power general matrix multiplication accelerator with sparse weight-and-output stationary dataflow,'' \emph{Micromachines}, vol.~16, no.~1, p. 101, 2025.

\bibitem{sparch}
Z.~Zhang, H.~Wang, S.~Han, and W.~J. Dally, ``Sparch: Efficient architecture for sparse matrix multiplication,'' in \emph{2020 IEEE International Symposium on High Performance Computer Architecture (HPCA)}.\hskip 1em plus 0.5em minus 0.4em\relax IEEE, 2020, pp. 261--274.

\bibitem{2023REMAP}
B.~Zhao, T.~Xia, H.~Zhai, F.~Ma, Y.~Du, H.~Chang, W.~Zhao, and P.~Ren, ``Remap: A spatiotemporal cnn accelerator optimization methodology and toolkit thereof,'' \emph{IEEE Transactions on Computer-Aided Design of Integrated Circuits and Systems: A publication of the IEEE Circuits and Systems Society}, no.~5, p.~42, 2023.

\bibitem{2020GAMMA}
T.~Krishna and S.~Kao, ``Gamma: automating the hw mapping of dnn models on accelerators via genetic algorithm,'' in \emph{ICCAD '20: IEEE/ACM International Conference on Computer-Aided Design}, 2020.

\bibitem{kao2020confuciux}
S.-C. Kao, G.~Jeong, and T.~Krishna, ``Confuciux: Autonomous hardware resource assignment for dnn accelerators using reinforcement learning,'' in \emph{2020 53rd Annual IEEE/ACM International Symposium on Microarchitecture (MICRO)}.\hskip 1em plus 0.5em minus 0.4em\relax IEEE, 2020, pp. 622--636.

\bibitem{ocf1}
I.~Mehrez, O.~Hamdi-Larbi, T.~Dufaud, and N.~Emad, ``Towards an auto-tuning system design for optimal sparse compression format selection with user expertise,'' in \emph{2016 IEEE/ACS 13th International Conference of Computer Systems and Applications (AICCSA)}.\hskip 1em plus 0.5em minus 0.4em\relax IEEE, 2016, pp. 1--6.

\bibitem{ocf2_cdpu}
S.~Karandikar, A.~N. Udipi, J.~Choi, J.~Whangbo, J.~Zhao, S.~Kanev, E.~Lim, J.~Alakuijala, V.~Madduri, Y.~S. Shao \emph{et~al.}, ``Cdpu: Co-designing compression and decompression processing units for hyperscale systems,'' in \emph{Proceedings of the 50th Annual International Symposium on Computer Architecture}, 2023, pp. 1--17.

\bibitem{ocf3_extending}
E.~Qin, G.~Jeong, W.~Won, S.-C. Kao, H.~Kwon, S.~Srinivasan, D.~Das, G.~E. Moon, S.~Rajamanickam, and T.~Krishna, ``Extending sparse tensor accelerators to support multiple compression formats,'' in \emph{2021 IEEE International Parallel and Distributed Processing Symposium (IPDPS)}.\hskip 1em plus 0.5em minus 0.4em\relax IEEE, 2021, pp. 1014--1024.

\bibitem{mapping_dse}
E.~Russo, M.~Palesi, S.~Monteleone, D.~Patti, G.~Ascia, and V.~Catania, ``Medea: A multi-objective evolutionary approach to dnn hardware mapping,'' in \emph{2022 Design, Automation \& Test in Europe Conference \& Exhibition (DATE)}.\hskip 1em plus 0.5em minus 0.4em\relax IEEE, 2022, pp. 226--231.

\bibitem{sparse_reward1}
M.~Riedmiller, R.~Hafner, T.~Lampe, M.~Neunert, J.~Degrave, T.~Wiele, V.~Mnih, N.~Heess, and J.~T. Springenberg, ``Learning by playing solving sparse reward tasks from scratch,'' in \emph{International conference on machine learning}.\hskip 1em plus 0.5em minus 0.4em\relax PMLR, 2018, pp. 4344--4353.

\bibitem{sparse_reward2}
J.~Hare, ``Dealing with sparse rewards in reinforcement learning,'' \emph{arXiv preprint arXiv:1910.09281}, 2019.

\bibitem{dstc}
Y.~Wang, C.~Zhang, Z.~Xie, C.~Guo, Y.~Liu, and J.~Leng, ``Dual-side sparse tensor core,'' in \emph{2021 ACM/IEEE 48th Annual International Symposium on Computer Architecture (ISCA)}.\hskip 1em plus 0.5em minus 0.4em\relax IEEE, 2021, pp. 1083--1095.

\bibitem{rle_cite}
S.~B. Wright, ``An overview of data compression techniques,'' Ph.D. dissertation, University of Washington, 1989.

\bibitem{deepbench}
\BIBentryALTinterwordspacing
(2016). [Online]. Available: \url{https://github.com/baidu-research/deepbench}
\BIBentrySTDinterwordspacing

\bibitem{pso_dse}
A.~Sengupta and V.~K. Mishra, ``Integrated particle swarm optimization (i-pso): An adaptive design space exploration framework for power-performance tradeoff in architectural synthesis,'' in \emph{Fifteenth International Symposium on Quality Electronic Design}.\hskip 1em plus 0.5em minus 0.4em\relax IEEE, 2014, pp. 60--67.

\bibitem{gebrekidan2024autonomous}
S.~B. Gebrekidan and S.~Marburg, ``Autonomous design of noise-mitigating structures using deep reinforcement learning,'' \emph{The Journal of the Acoustical Society of America}, vol. 156, no.~1, pp. 151--163, 2024.

\bibitem{es1}
K.~Deb, A.~Pratap, S.~Agarwal, and T.~Meyarivan, ``A fast and elitist multiobjective genetic algorithm: Nsga-ii,'' \emph{IEEE transactions on evolutionary computation}, vol.~6, no.~2, pp. 182--197, 2002.

\bibitem{es2}
G.~M. Morris, D.~S. Goodsell, R.~S. Halliday, R.~Huey, W.~E. Hart, R.~K. Belew, and A.~J. Olson, ``Automated docking using a lamarckian genetic algorithm and an empirical binding free energy function,'' \emph{Journal of computational chemistry}, vol.~19, no.~14, pp. 1639--1662, 1998.

\bibitem{es3}
G.~Jones, P.~Willett, R.~C. Glen, A.~R. Leach, and R.~Taylor, ``Development and validation of a genetic algorithm for flexible docking,'' \emph{Journal of molecular biology}, vol. 267, no.~3, pp. 727--748, 1997.

\bibitem{es4}
J.~Schaffer, ``Multiobjective optimization using nondominated sorting in genetic algorithms,'' in \emph{Proceedings of the First International Conference on Genetic Algorithms and Their Applications}.\hskip 1em plus 0.5em minus 0.4em\relax Lawrence Erlbaum Associates, 1985, pp. 160--168.

\bibitem{ga_corelative1}
G.~Zames, ``Genetic algorithms in search, optimization and machine learning,'' \emph{Inf Tech J}, vol.~3, no.~1, p. 301, 1981.

\bibitem{ga_corelative2}
M.~Mitchell, \emph{An introduction to genetic algorithms}.\hskip 1em plus 0.5em minus 0.4em\relax MIT press, 1998.

\bibitem{ga_corelative3}
J.~Zhong, L.~Feng, and Y.-S. Ong, ``Gene expression programming: A survey,'' \emph{IEEE Computational Intelligence Magazine}, vol.~12, no.~3, pp. 54--72, 2017.

\bibitem{rand_init1}
B.~Kazimipour, X.~Li, and A.~K. Qin, ``A review of population initialization techniques for evolutionary algorithms,'' in \emph{2014 IEEE congress on evolutionary computation (CEC)}.\hskip 1em plus 0.5em minus 0.4em\relax IEEE, 2014, pp. 2585--2592.

\bibitem{rand_init2}
E.~K. Burke, J.~P. Newall, and R.~F. Weare, ``Initialization strategies and diversity in evolutionary timetabling,'' \emph{Evolutionary computation}, vol.~6, no.~1, pp. 81--103, 1998.

\bibitem{rand_init3}
A.~Anand, M.~Degroote, and A.~Aspuru-Guzik, ``Natural evolutionary strategies for variational quantum computation,'' \emph{Machine Learning: Science and Technology}, vol.~2, no.~4, p. 045012, 2021.

\bibitem{cube_init1}
H.~Escobar-Cuevas, E.~Cuevas, K.~Avila, and O.~Avalos, ``An advanced initialization technique for metaheuristic optimization: a fusion of latin hypercube sampling and evolutionary behaviors,'' \emph{Computational and Applied Mathematics}, vol.~43, no.~4, p. 234, 2024.

\bibitem{cube_init2}
H.~R. Medeiros, D.~M. Izidio, A.~P. d.~A. Ferreira, and E.~N. da~S.~Barros, ``Latin hypercube initialization strategy for design space exploration of deep neural network architectures,'' in \emph{Proceedings of the Genetic and Evolutionary Computation Conference Companion}, 2019, pp. 295--296.

\bibitem{cube_init3}
M.~Hamdan and O.~Qudah, ``The initialization of evolutionary multi-objective optimization algorithms,'' in \emph{Advances in Swarm and Computational Intelligence: 6th International Conference, ICSI 2015, held in conjunction with the Second BRICS Congress, CCI 2015, Beijing, China, June 25-28, 2015, Proceedings, Part I 6}.\hskip 1em plus 0.5em minus 0.4em\relax Springer, 2015, pp. 495--504.

\bibitem{cube_init4}
H.~Chang, Y.~Sun, S.~Lu, and D.~Lin, ``A multistrategy differential evolution algorithm combined with latin hypercube sampling applied to a brain--computer interface to improve the effect of node displacement,'' \emph{Scientific Reports}, vol.~14, no.~1, p. 20420, 2024.

\bibitem{montecarlo}
R.~Y. Rubinstein and D.~P. Kroese, \emph{Simulation and the Monte Carlo method}.\hskip 1em plus 0.5em minus 0.4em\relax John Wiley \& Sons, 2016.

\bibitem{tpu}
N.~P. Jouppi, C.~Young, N.~Patil, D.~Patterson, G.~Agrawal, R.~Bajwa, S.~Bates, S.~Bhatia, N.~Boden, A.~Borchers \emph{et~al.}, ``In-datacenter performance analysis of a tensor processing unit,'' in \emph{Proceedings of the 44th annual international symposium on computer architecture}, 2017, pp. 1--12.

\bibitem{frantar2023sparsegpt}
E.~Frantar and D.~Alistarh, ``Sparsegpt: Massive language models can be accurately pruned in one-shot,'' in \emph{International Conference on Machine Learning}.\hskip 1em plus 0.5em minus 0.4em\relax PMLR, 2023, pp. 10\,323--10\,337.

\end{thebibliography}

\vfill

\end{document}